\pdfoutput=1

\documentclass[11pt]{article}

\usepackage[preprint]{acl}
\usepackage{balance}
\usepackage{times}
\usepackage{latexsym}
\usepackage{xspace}
\usepackage{devanagari}
\usepackage{booktabs}
\usepackage{array}
\usepackage{longtable}
\usepackage{amsmath}
\usepackage{enumitem}
\usepackage{hyperref}

\usepackage[T1]{fontenc}

\usepackage[utf8]{inputenc}

\usepackage{microtype}
\usepackage{textcase}
\newcommand{\newtextsc}[1]{{\small \textsf{#1}}}

\newcommand{\evaluator}{\newtextsc{Qwen-2.5-72B-Inst}\xspace}
\newcommand{\factual}{\newtextsc{Factual Recall}\xspace}
\newcommand{\incontext}{\newtextsc{In-context Recall}\xspace}
\newcommand{\incontextrobust}{\newtextsc{Counter-Factual Context Adherence}\xspace}

\newcommand{\orca}{\newtextsc{Orca-2-7B}\xspace}

\newcommand{\deepseek}[1]{\newtextsc{DeepSeek-#1}\xspace}

\newcommand{\mistralins}[1]{\newtextsc{Mistral-#1-v0.2}\xspace}

\newcommand{\metallamains}[1]{\newtextsc{Llama-3-#1}\xspace}
\newcommand{\llamains}[1]{\newtextsc{Llama-3.2-#1}\xspace}
\newcommand{\phimodel}[1]{\newtextsc{Phi-#1}\xspace}
\newcommand{\gemmains}[1]{\newtextsc{Gemma-2-#1}\xspace}

\newcommand{\dbrx}[1]{\newtextsc{DBRX-Base}\xspace}

\usepackage{inconsolata}

\usepackage{graphicx}

\title{Language Models’ Factuality Depends on the Language of Inquiry}

\usepackage{xcolor}
\usepackage{amssymb}
\usepackage{hyperref}

\newcommand{\diamondshape}{\ensuremath{\diamondsuit}}
\newcommand{\spade}{\ensuremath{\spadesuit}}
\newcommand{\boxshape}{\ensuremath{\Box}}
\newcommand{\triangleshape}{\ensuremath{\triangle}}

\newcommand{\starshape}{\ensuremath{\bigstar}}
\newcommand{\blackclub}{\textcolor{black}{\ensuremath{\clubsuit}}}
\newcommand{\blackheart}{\textcolor{black}{\ensuremath{\heartsuit}}}
\title{Language Models' Factuality Depends on the Language of Inquiry}

\author{
  \textbf{Tushar Aggarwal}$^{\textbf{*},\blackheart}$,
  \textbf{Kumar Tanmay}$^{\textbf{*},\spade,\starshape}$,
  \textbf{Ayush Agrawal}$^{\textbf{*},\blackheart,\blackclub,\diamondshape}$,
  \textbf{Kumar Ayush}$^{\boxshape,\triangleshape}$,
  \\
  \textbf{Hamid Palangi}$^{\triangleshape}$,
  \textbf{Paul Pu Liang}$^{\starshape}$ \\
  $^{\spade}$Harvard University, \ 
  $^{\blackclub}$Universit\'e de Montr\'eal, \ 
  $^{\diamondshape}$Mila, \ 
  $^{\starshape}$MIT\\
  $^{\boxshape}$Stanford University, \ 
  $^{\triangleshape}$Google, \ 
  $^{\blackheart}$Microsoft Research\\
}

\begin{document}

\maketitle

\begin{abstract}

Multilingual language models (LMs) are expected to recall factual knowledge consistently across languages, yet they often fail to transfer knowledge between languages even when they possess the correct information in one of the languages. For example, we find that an LM may correctly identify \textit{Rashed Al Shashai} as being from \textit{Saudi Arabia} when asked in Arabic, but consistently fails to do so when asked in English or Swahili. To systematically investigate this limitation, we introduce a benchmark of 10,000 country-related facts across 13 languages and propose three novel metrics—Factual Recall Score, Knowledge Transferability Score, and Cross-Lingual Factual Knowledge Transferability Score—to quantify factual recall and knowledge transferability in LMs across different languages. Our results reveal fundamental weaknesses in today's state-of-the-art LMs, particularly in cross-lingual generalization where models fail to transfer knowledge effectively across different languages, leading to inconsistent performance sensitive to the language used. Our findings emphasize the need for LMs to recognize language-specific factual reliability and leverage the most trustworthy information across languages. We release our benchmark and evaluation framework to drive future research in multilingual knowledge transfer. The data and codes are available at this \href{https://github.com/kmrtanmay/X_FaKT.git}{link}.
\end{abstract}
\renewcommand{\thefootnote}{}
\footnotemark\footnotetext{*equal contribution.}
\footnotetext{Corresponding authors: tushar.aggarwal53@gmail.com, kumartanmay@fas.harvard.edu, ayush.agrawal@mila.quebec}

\section{Introduction}

\begin{figure}[t]  
    \centering
    \includegraphics[width=1.\linewidth, height=5cm]{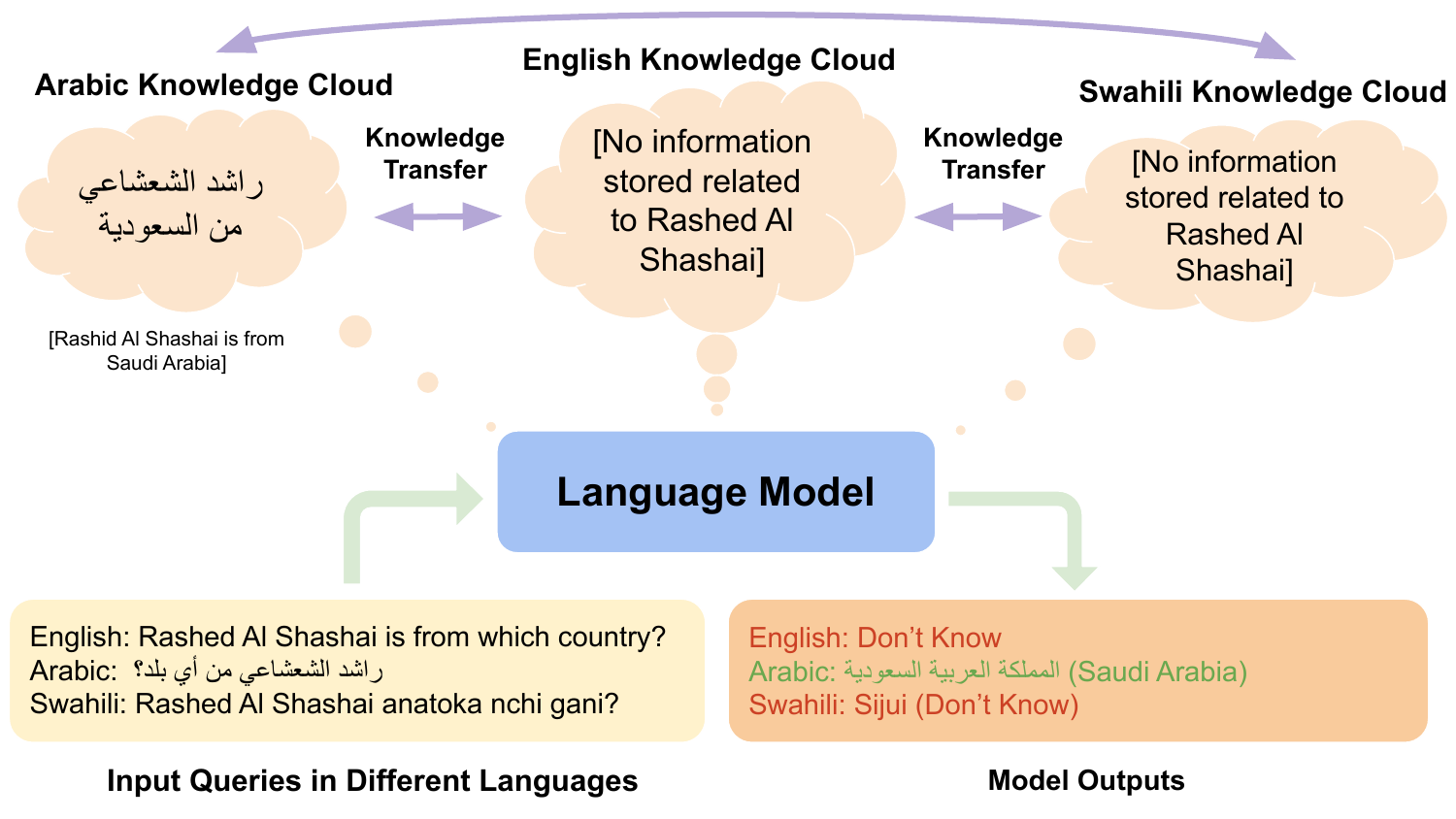}  
    \caption{Illustratation of the cross-lingual factual knowledge transferability issue across linguistic knowledge clouds in LMs. The model correctly recalls that \textit{Rashed Al Shashai is from Saudi Arabia} when queried in Arabic, but fails to retrieve this fact in English and Swahili, highlighting that factual knowledge is often stored in language-specific silos.}
    \label{fig:knowledge_transfer}  
\end{figure}

Large Language Models (LLMs) are often perceived as vast knowledge reservoirs, capable of recalling factual information across multiple languages \cite{wang-etal-2024-knowledge-mechanisms}. However, what if their knowledge is locked within linguistic boundaries and unable to be transferred across languages? Despite advancements in multilingual LMs such as Llama \cite{touvron2023llama, dubey2024llama}, Gemma \cite{team2024gemma}, DeepSeek \cite{deepseekai2024deepseekllmscalingopensource}, and Phi \cite{abdin2024phi, li2023textbooks}, our study reveals a striking asymmetry in their factual recall across languages: consider the example in Figure~\ref{fig:knowledge_transfer}, where an LM is tasked with a simple factual query: ``\textit{Rashed Al Shashai is from which country?}'' When asked in Arabic, several state-of-the-art LMs correctly generate the response: ``\textit{Saudi Arabia}.'' However, when posed in English, Hindi, or Swahili, the same models fail to recall the fact.
This example suggests that models can correctly retrieve country-specific facts in the language associated with that country but struggle to do so in others.
 
This raises a critical question—do these models truly internalize and transfer factual knowledge across languages, or do they merely encode isolated linguistic silos?

This limitation has significant implications for multilingual AI development and real-world applications. Many LM-based systems—such as retrieval-augmented generation (RAG) pipelines, multilingual search engines, and cross-lingual reasoning models—assume that factual knowledge is consistently available and transferable across languages. 

Our findings reveal that LMs often rely on language-specific memorization rather than true cross-lingual knowledge generalization. This over-reliance can introduce biases, inconsistencies, and reliability issues in multilingual AI applications~\cite{chua2024crosslingual}. 

To systematically analyze the factual inconsistencies,  we introduce a carefully curated dataset comprising country-related facts translated into 13 languages. This benchmark evaluates LMs on multiple dimensions—\textit{factual recall, in-context recall, and counter-factual context adherence}—across high-, medium-, and low-resource languages. This benchmark comprises of 802 instances for factual recall, 156 instances for In-context recall, and 1404 instances for counter-factual context adherence as shown in Table~\ref{tab:data_stats}.

\textit{Factual recall} assesses the LM's ability to recall country-specific facts consistently across multiple languages. We evaluate factual recall using three metrics: (a) \textit{Factual Recall Score (FRS)}: Measures how accurately a model recalls a fact in a given language, (b) \textit{Knowledge Transferability Score (KTS)}: Quantifies how well factual knowledge is transferred across languages, and (c) \textit{Cross-Lingual Factual Knowledge Transferability (X-FaKT) Score}: Combines the assessment of factual recall and cross-lingual transfer ability. FRS and KTS measure the effectiveness of cross-lingual knowledge transfer, and X-FaKT Score integrates factual recall with transferability to provide a robust measure of multilingual generalization. These metrics offer a more nuanced evaluation than a simple error rate, allowing for a deeper understanding of cross-lingual generalization.

\textit{In-Context Recall}~\cite{machlab2024llmincontextrecallprompt} measures the general performance of the models in multilingual contexts. Inspired by \cite{cotterellcontext}, we also study how factual knowledge of models affects their performance in handling in-context tasks in the multilingual setting (\textit{Counterfactual Context Adherence}). For this, we design a dataset where factual knowledge conflicts with in-context instructions.

Our experiments reveal that while LMs often retrieve factual information correctly in the language associated with the fact, they struggle to transfer this knowledge to other languages. We also found that the size of the LLM plays an important role in factuality and knowledge transferability. For example, the combined performance of LLama-3-70B in factuality and knowledge transfer across languages is markedly (~152\% $\uparrow$ in \textit{X-FaKT} Score) better than Llama-3.2-1B. In addition, there is a marked difference in these tasks when queries are asked in high-resource languages (~46\% $\uparrow$ in \textit{X-FaKT} Score) as compared to the case with low resources. This finding exposes a critical limitation in current language models and their approach to multilingual knowledge integration. Our findings also reveal an interesting trade-off: LMs with stronger factual recall often struggle with counterfactual adherence, highlighting a key limitation in balancing factual memory and contextual reasoning. 
In our experiments, we observed that the factual knowledge of LMs could skew their judgments, leading to inaccurate evaluations. One has to be very careful when designing the prompt and using LM as an evaluator. We highlight the importance of controlling the evaluator's factual knowledge to ensure consistent and effective evaluation.

\section{Related Work}

\textbf{Multilingual Transformers.} Early work by \cite{petroni-etal-2019-language} explored whether LMs can store factual knowledge about entities, setting the stage for later investigations into multilingual LMs. Notable multilingual models such as mBERT \cite{devlin-etal-2019-bert}, XLM-R \cite{conneau-etal-2020-unsupervised}, mT5 \cite{xue-etal-2021-mt5}, and BLOOM \cite{workshop2023bloom176bparameteropenaccessmultilingual} have demonstrated varying levels of performance across different languages. These models, trained on diverse multilingual corpora, show that LMs exhibit language-dependent capabilities in factual recall. Research has highlighted systematic biases in factual retrieval across languages \cite{artetxe-etal-2020-cross, liu-etal-2020-multilingual-denoising, kassner-etal-2021-multilingual}, which is a key challenge in multilingual LMs. While multilingual QA benchmarks like XQuAD \cite{artetxe-etal-2020-cross}, MLQA \cite{lewis-etal-2020-mlqa}, and TyDiQA \cite{clark-etal-2020-tydi} assess factual consistency, they do not directly measure the transfer of knowledge between languages. Recent work by \cite{wang-etal-2024-knowledge-mechanisms} raised questions about LMs' ability to recall factual knowledge in reasoning tasks, while \cite{fierro2025multilinguallanguagemodelsremember} emphasized the need for more robust methodologies for evaluating knowledge in multilingual LMs. Our study builds on these insights by introducing a benchmark specifically designed to assess cross-lingual factual knowledge transferability.

\textbf{Cross-Lingual Knowledge Transfer in LMs.} Recent works have sought to understand the factors that influence cross-lingual knowledge transfer in multilingual models. Studies suggest that multilingual LMs exhibit zero-shot and few-shot generalization across languages \cite{nooralahzadeh-etal-2020-zero, pfeiffer-etal-2020-mad}, but empirical evidence indicates that this transfer is often asymmetric, with high-resource languages benefiting more than lower-resource ones \cite{hu2020xtrememassivelymultilingualmultitask}. \cite{muller-etal-2021-first} investigated the connection between cross-lingual similarity in hidden representations and downstream task performance, revealing that LMs with stronger representation alignment across languages perform better. \cite{chai-etal-2022-cross} explored cross-linguality from a language structure perspective, emphasizing the importance of compositional properties in facilitating knowledge transfer. More recent work has focused on cross-lingual transfer from high-resource to low-resource languages \cite{zhao2024llama, zhao2024tracing}, further underscoring the asymmetries in cross-lingual knowledge integration. Our work contributes to this area by evaluating the effectiveness of factual knowledge transfer across languages using a comprehensive set of metrics designed to measure both factual recall and transferability.

\textbf{Context Sensitivity and Counterfactual Reasoning.} LMs are known to be highly sensitive to contextual cues, which can sometimes override factual knowledge when the context is misleading \cite{brown2020language, tirumala2022memorization, cotterellcontext}. \cite{ghosh2025multilingualmindsurvey} provides an in-depth review of multilingual reasoning in LMs. Counterfactual reasoning, in which models must consider hypothetical situations, has been studied in various contexts \cite{wu2023reasoning}. These studies show that LMs optimized for factual recall often struggle with counterfactual tasks, especially when faced with conflicting contextual instructions. While most prior evaluations have focused on monolingual settings \cite{shwartz-etal-2020-unsupervised, wang2020language}, our work extends these investigations into the multilingual domain. By introducing tasks like in-context recall and counterfactual adherence, we analyze how multilingual models handle both factual accuracy and contextual reasoning across languages, revealing important challenges in balancing factual knowledge and context sensitivity.

\section{Dataset}
\label{sec:datasets}

\begin{table}[t]
\centering
\begin{tabular}{@{}c|c@{}}
\toprule
\textbf{Task Type}                 &     \textbf{\# Examples}\\ \midrule
\factual & 802 \\
\incontext & 156 \\
\incontextrobust & 1404
 \\ \bottomrule
\end{tabular}
\caption{Number of examples per languages in our benchmark (\S\ref{sec:datasets}).}
\label{tab:data_stats}
\end{table}

\begin{figure}[t]  
    \centering
    \includegraphics[width=1.\linewidth]{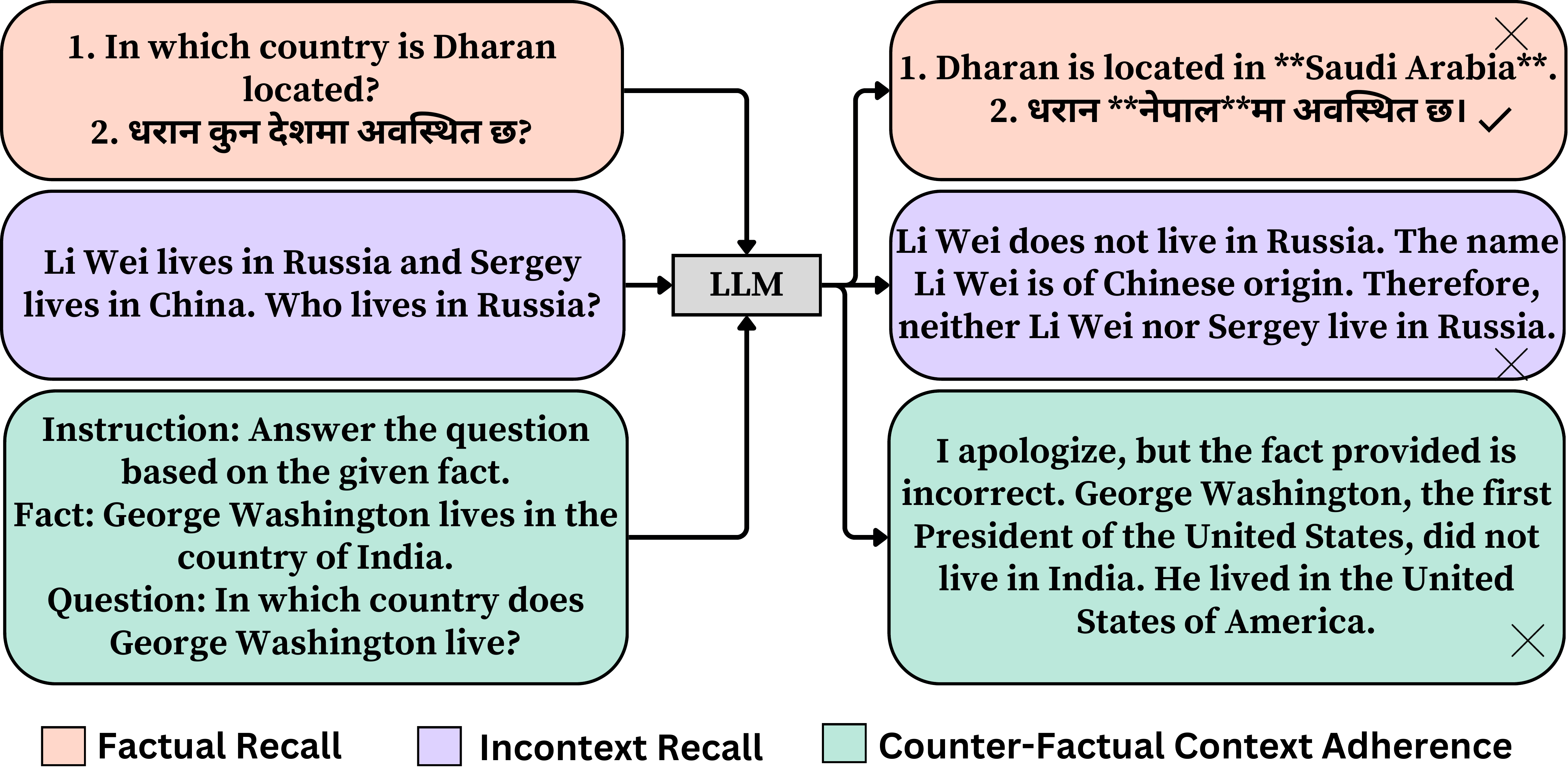}  
    \caption{Examples from our multilingual dataset illustrating three tasks. Factual Recall: LMs recall country-specific facts better in native languages, as seen with Dharan's correct identification in Nepali but incorrect in English. Incontext Recall: Models struggle with contextual reasoning, showing regional bias when associating names with countries. Counter-Factual Context Adherence: When given counterfactual prompts about well-known figures, models rely on prior knowledge, affecting their ability to adhere to provided context. } 
    \label{fig:examples}  
\end{figure}

We introduce a new multilingual dataset designed to evaluate three key capabilities of LMs: (a) \textit{Factual Recall}, (b) \textit{In-context Recall}, and (c) \textit{Counter-Factual Context Adherence}. The number of instances in our dataset is given in the Table~\ref{tab:data_stats}. Given the multilingual nature of our study, we categorize languages based on their resource availability in existing LM training corpora:

\textbf{High-resource}: English, Chinese, French, Japanese.

\textbf{Medium-resource}: Hindi, Russian, Arabic, Greek.

\textbf{Low-resource}: Nepali, Ukrainian, Turkish, Swahili, Thai.

These languages correspond to countries strongly associated with their usage: the United States, China, France, Japan, India, Russia, Saudi Arabia, Greece, Nepal, Ukraine, Turkey, Kenya, and Thailand. Now, we describe our datasets in detail.

\subsection{Factual Recall}

This task evaluates an LM’s ability to recall country-specific facts across multiple languages. For example, given the query, \textit{In which country is Mumbai located?}, the model should correctly respond with \textit{India} when asked in different languages.

To construct the dataset, we curated a diverse set of entities—including cities, artists, sports figures, landmarks, festivals, and politicians—for 13 selected countries. We then created standardized templates for factual queries and translated them into each language using the Google Translate API~\cite{google_translate}. All translations were manually verified and refined as needed with the assistance of ChatGPT. In total, our dataset consists of 805 unique factual questions, each available in 13 language versions.

\subsection{In-Context Recall}

The in-context recall task evaluates how effectively an LM utilizes contextual information to answer a question, ensuring that internal knowledge does not influence the model’s output.

Building on the work of \cite{feng2024languagemodelsbindentities}, we constructed our dataset by focusing on common person names associated with each country. For each example, we sampled two names and paired them with two different countries, creating context-based prompts as shown in violet color in Figure~\ref{fig:examples}.  To enhance dataset efficiency, we intentionally avoided associating a name with its most commonly linked country within the example.

\subsection{\textbf{Counter-Factual Context Adherence}}

This task evaluates an LM’s susceptibility to counterfactual information by assessing whether it adheres to the provided context when answering a question. Ideally, the model should rely solely on the given context, but in some cases, its internal knowledge may interfere or override it, leading to unintended responses~\cite{cotterellcontext}. To investigate this, we curated a list of well-known personalities strongly associated with specific countries and deliberately introduced counterfactual information into the context.

For the example given in Figure~\ref{fig:examples}, if the model defaults to its internal knowledge and answers \textit{United States}, it demonstrates a resistance to the contextual information. Conversely, if it follows the counterfactual context and answers \textit{India}, it suggests a higher reliance on the provided context rather than pre-existing knowledge.

One might expect these models to perform near-perfectly on these tasks, as they are very simple. However, despite the simplicity of these tasks, the performance varies across languages and models.

\section{Experiments}

In this section, we discuss our experimental setup, metric formulation, and both quantitative and qualitative analyses. We present the results of our experiments evaluating LMs on our dataset across diverse multilingual tasks. These experiments assess how language and country-specific factual knowledge influence LMs responses in a multilingual setting. All experiments were conducted using the latest models, with \evaluator~\cite{qwen2025qwen25technicalreport} serving as the evaluator \cite{li2024llmsasjudgescomprehensivesurveyllmbased}.
 
\subsection{Experimental Setup}
\paragraph{Models} We evaluated 14 models of varying sizes, trained on different compositions of multilingual data, and fine-tuned using various preference optimization strategies~\cite{ouyang2022training, rafailov2024directpreferenceoptimizationlanguage}, for our multilingual study. These include Deepseek~\cite{deepseekai2024deepseekllmscalingopensource}, Qwen~\cite{qwen2}, Gemma~\cite{gemmateam2024gemmaopenmodelsbased}, and Llama~\cite{touvron2023llamaopenefficientfoundation} families. Further details of the models evaluated are given in Table~\ref{tab:model-specs}.

\paragraph{Compute Details}
All our experiments were conducted on a set of 4 NVIDIA A100 GPUs, each with 80GB of VRAM. 

\paragraph{Evaluation}
To evaluate all models on the curated datasets (Section~\ref{sec:datasets}), we used a temperature setting of 0 and a maximum token limit of 128. Specifically, we tested the models' performance on Factual Recall and In-Context Recall across different settings. For evaluation, we designed our metrics and utilized \evaluator as the evaluator \cite{li2024llmsasjudgescomprehensivesurveyllmbased}, with a maximum token limit of 256 to support reasoning. Evaluation prompts are shown in Figures~\ref{fig:prompt1} and~\ref{fig:prompt2}.

\subsection{Metric Definition and Formulation}
This section introduces our carefully designed metrics to evaluate factual recall and knowledge transferability across languages in LMs. We propose two key metrics: the \textit{Factual Recall Score (FRS)} and the \textit{Knowledge Transferability Score (KTS)}.
To establish a common metric for evaluating the model's performance in our benchmark, we compute their harmonic mean, which is defined as the \textit{Cross-Lingual Factual Knowledge Transferability Score (X-FaKT)}, to ensure a balanced assessment while penalizing large disparities between them. Our metrics incorporate an inverse formulation with a correction factor to maintain a bounded range of $[0,1]$. A higher error rate results in a lower metric value due to the inverse transformation, ensuring that better model performance corresponds to higher scores.

\subsubsection{Associative vs. Non-Associative Knowledge}  
We categorize our dataset into two groups: associative and non-associative knowledge. The categorization is defined as follows: we consider 13 languages, each associated with a corresponding country (i.e., the $i$th language belongs to the $i$th country).

\vspace{1em}
\text{Associative} = $\{Q \in \text{Questions} : Q \in \text{Language}_i \wedge \text{output}(Q) = \text{Country}_j \wedge  i = j\}$

\vspace{1em}

\text{Non-associative} = $\{Q \in \text{Questions} : Q \in \text{Language}_i \wedge \text{output}(Q) = \text{Country}_j \wedge  i \neq j\}$

\vspace{1em}
We denote the mean error rate for a country-specific fact asked in the language strongly associated with that country as $ \mu_{\text{assoc.}} $, and the mean error rate for a country-specific fact asked in a language not associated with that country as $ \mu_{\text{non-assoc.}} $.

\subsubsection{Factual Recall Score (FRS)}
Factual recall evaluates the model's ability to correctly retrieve both \textit{associative} and \textit{non-associative} knowledge. We define the Factual Recall Score (FRS) as:

\vspace{-15pt} 
\begin{equation}
FRS = \frac{3}{2} \left( \frac{1}{\mu_{\text{assoc.}} + \mu_{\text{non-assoc.}} + 1} - \frac{1}{3} \right)
\end{equation}
\vspace{-15pt} 

\begin{itemize}
    \setlength{\itemsep}{0pt} 
    \setlength{\parskip}{0pt} 
    \setlength{\parsep}{0pt}  
    \item When both errors are zero ($ \mu_{\text{assoc.}} = 0, \mu_{\text{non-assoc.}} = 0 $), the model has a perfect factual recall, yielding an FRS score of 1.
    \item When both errors are high, the denominator increases, resulting in a lower FRS score closer to 0, indicating poor factual recall.  
\end{itemize}

\subsubsection{Knowledge Transferability Score (KTS)}
Knowledge transferability quantifies how well a model maintains consistent factual knowledge across languages. We define the \textit{Knowledge Transferability Score (KTS)} as:

\vspace{-15pt} 
\begin{equation}
KTS = 2 \left( \frac{1}{\left|\mu_{\text{assoc.}} - \mu_{\text{non-assoc.}} \right| + 1} - \frac{1}{2} \right)
\end{equation}
\vspace{-15pt} 

where:
\begin{itemize}
    \setlength{\itemsep}{0pt} 
    \setlength{\parskip}{0pt} 
    \setlength{\parsep}{0pt}  
    \item $ |\mu_{\text{assoc.}} - \mu_{\text{non-assoc.}}| $ captures the absolute difference between associative and non-associative recall errors.
\end{itemize}

\begin{itemize}
    \setlength{\itemsep}{0pt} 
    \setlength{\parskip}{0pt} 
    \setlength{\parsep}{0pt}  
    \item When both errors are zero ($ \mu_{\text{assoc.}} = 0, \mu_{\text{non-assoc.}} = 0 $), there is perfect factual knowledge transfer, resulting in a KTS score of 1.
    \item When both errors are high but equal (e.g., $ \mu_{\text{assoc.}} = 20, \mu_{\text{non-assoc.}} = 20 $), KTS remains 1, indicating that while factual recall is poor, the model exhibits consistent errors across languages.
    \item When errors differ significantly (e.g., $ \mu_{\text{assoc.}} = 20, \mu_{\text{non-assoc.}} = 2 $ or vice versa), the absolute difference increases, leading to a lower KTS, highlighting a lack of knowledge transfer across languages.
\end{itemize}

\subsubsection{Cross-Lingual Factual Knowledge Transferability Score (X-FAKT)}
 To ensure a balanced evaluation of factual recall and cross-lingual transferability, we compute their harmonic mean:

\begin{equation}
X\text{-FAKT} = 2 \times \frac{FRS \times KTS}{FRS + KTS}
\end{equation}

where:
\begin{itemize}
    \setlength{\itemsep}{0pt} 
    \setlength{\parskip}{0pt} 
    \setlength{\parsep}{0pt}  
    \item The harmonic mean penalizes large disparities between factual recall (FRS) and knowledge transferability (KTS), ensuring that both contribute meaningfully to the final score.
    \item If either FRS or KTS is significantly lower, the overall score remains low, discouraging models from excelling in one metric while performing poorly in the other.
    \item A high X-FAKT score indicates that the model is both factually accurate and consistent across multiple languages.
\end{itemize}

This formulation provides a holistic evaluation of factual knowledge retention and cross-lingual consistency, making it a robust metric for assessing multilingual model performance.

\subsection{Quantitative Analysis}
\subsubsection{Performance on Factual Recall task}

\begin{figure}[t]  
    \centering
    \includegraphics[width=1\linewidth]{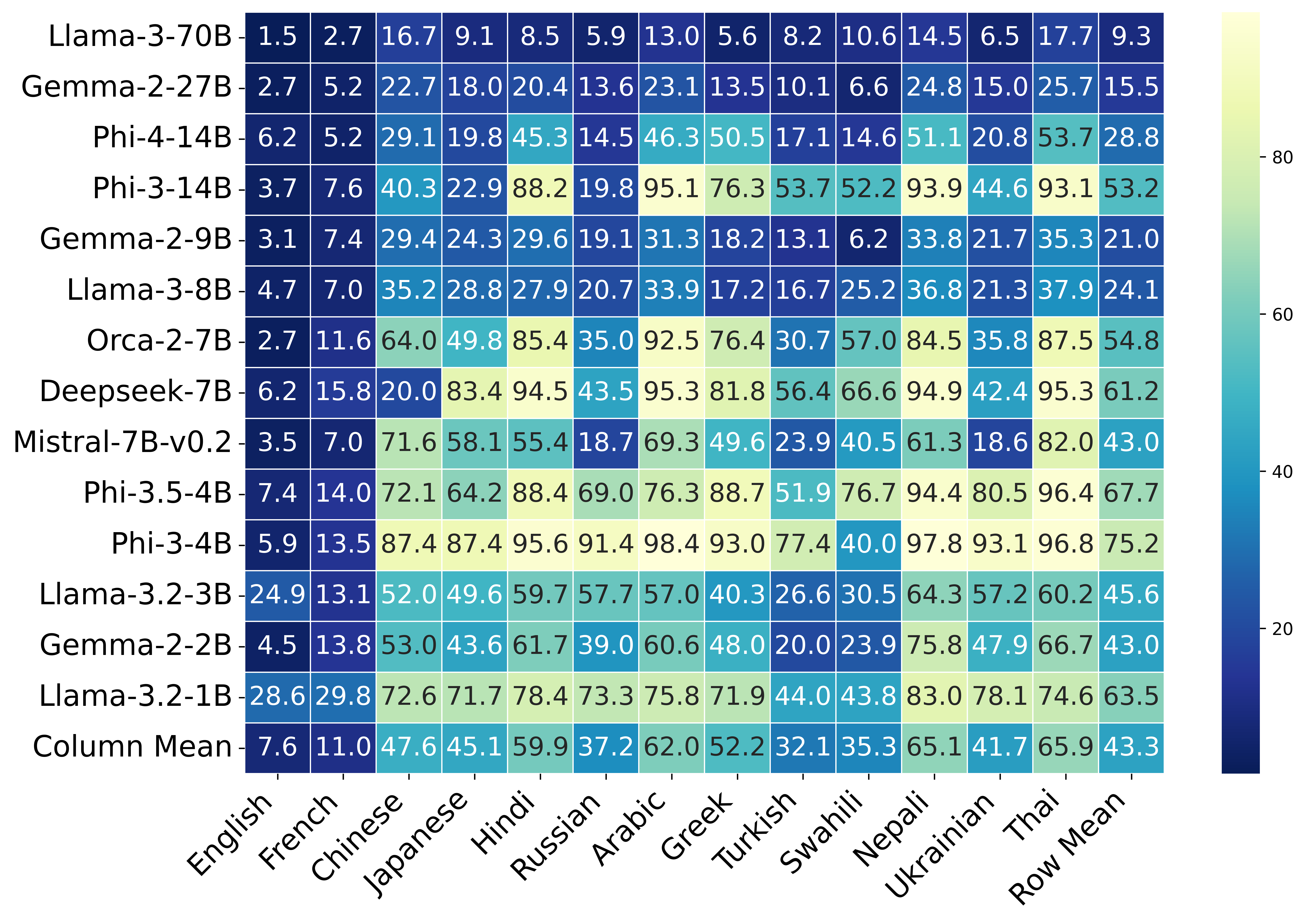}  
    \caption{Error rates for each model on the Factual Recall task. A clear pattern emerges, showing a decline in performance as we move from larger to smaller models (top to bottom) and from high-resource to low-resource languages (left to right).}
    \label{fig:factual_recall}  
\end{figure}

The error rate across different LMs (Figure~\ref{fig:factual_recall}) reveals a clear pattern in performance across languages and model sizes. Notably, all models demonstrate superior performance on high-resource languages like English and French, with error rates consistently below 15\% for most model variants. This performance gradually deteriorates as the model size decreases, with smaller models showing significantly higher error rates across all languages. However, an interesting observation emerges with languages like Swahili and Turkish, which despite being low-resource languages, exhibit relatively better performance with error rates comparable to mid-resource languages. This can be attributed to their use of Latin script, facilitating better knowledge transfer from English.

A compelling pattern emerges when examining languages that share similar scripts, and strong correlations in model performance among languages that share similar scripts. For example, the error patterns for Hindi-Nepali and  Russian-Ukrainian pairs show remarkable similarities, suggesting that the models effectively leverage shared scriptural characteristics during learning. These patterns indicate that script similarity plays a crucial role in the model's ability to generalize across languages, potentially offering insights into how these models transfer knowledge between different language pairs and scripts.

\begin{table}[t]
\centering
\small
\resizebox{0.49\textwidth}{!}{
\begin{tabular}{l|c|c|c|c|c|c|c}
\textbf{Model} & $\boldsymbol{\mu_{assoc.} (\%)}$ & $\boldsymbol{\mu_{non-assoc.} (\%)}$ & \textbf{t-stat} & \textbf{p-value} & \textbf{FRS} & \textbf{KTS} & \textbf{X-FAKT} \\
\midrule
\metallamains{70B} & \textbf{2.36} $\pm$ 5.12 & \textbf{9.85} $\pm$ 10.54 & 2.52 & 0.01 & \textbf{0.835} & \textbf{0.862} & \textbf{0.848} \\
\gemmains{27B} & 4.23 $\pm$ 8.49 & 16.46 $\pm$ 17.07 & 2.54 & 0.01 & 0.742 & 0.783 & 0.762 \\
\phimodel{4-14B} & 12.87 $\pm$ 16.51 & 30.15 $\pm$ 25.92 & 2.35 & 0.02 & 0.548 & 0.706 & 0.617 \\
\phimodel{3-14B} & 25.09 $\pm$ 29.84 & 55.57 $\pm$ 36.24 & 2.93 & $<$0.01 & 0.330 & 0.535 & 0.408 \\
\gemmains{9B} & 4.98 $\pm$ 6.09 & 22.32 $\pm$ 21.37 & 2.90 & $<$0.01 & 0.677 & 0.705 & 0.691 \\
\metallamains{8B} & 4.60 $\pm$ 7.54 & 25.77 $\pm$ 19.61 & 3.85 & $<$0.01 & 0.649 & 0.651 & 0.650 \\
\orca & 31.95 $\pm$ 31.65 & 56.77 $\pm$ 32.99 & 2.60 & 0.01 & 0.295 & 0.603 & 0.396 \\
\deepseek{7b} & 31.49 $\pm$ 30.68 & 63.73 $\pm$ 36.29 & 3.09 & $<$0.01 & 0.268 & 0.514 & 0.353 \\
\mistralins{7B} & 16.96 $\pm$ 15.65 & 45.25 $\pm$ 29.34 & 3.42 & $<$0.01 & 0.424 & 0.559 & 0.483 \\
\phimodel{3.5-4B} & 41.85 $\pm$ 31.62 & 69.87 $\pm$ 31.23 & 3.09 & $<$0.01 & 0.208 & 0.563 & 0.304 \\
\phimodel{3-4B} & 42.45 $\pm$ 30.99 & 77.95 $\pm$ 33.72 & 3.65 & $<$0.01 & 0.181 & 0.477 & 0.262 \\
\llamains{3B} & 24.10 $\pm$ 17.80 & 47.48 $\pm$ 26.80 & 3.07 & $<$0.01 & 0.375 & 0.620 & 0.467 \\
\gemmains{2B} & 9.97 $\pm$ 14.78 & 45.77 $\pm$ 31.30 & 4.06 & $<$0.01 & 0.463 & 0.473 & 0.468 \\
\llamains{1B} & 34.74 $\pm$ 22.32 & 65.96 $\pm$ 26.98 & 4.03 & $<$0.01 & 0.247 & 0.524 & 0.336 \\
\bottomrule
\end{tabular}
}
\caption{Results of the t-test comparing associative and non-associative knowledge across models, alongside FRS, KTS, and X-FAKT scores. (A) Llama-3-70B achieves the best performance in both factual recall and knowledge transferability. (B) There is a statistically significant difference between the performance on associative queries (asked in a country's native language) and non-associative queries (asked in other languages).}
\label{tab:tstats_factual}
\end{table}

\paragraph{Knowledge Transferability Analysis:}

\begin{figure}[t]  
    \centering
    \includegraphics[width=1\linewidth]{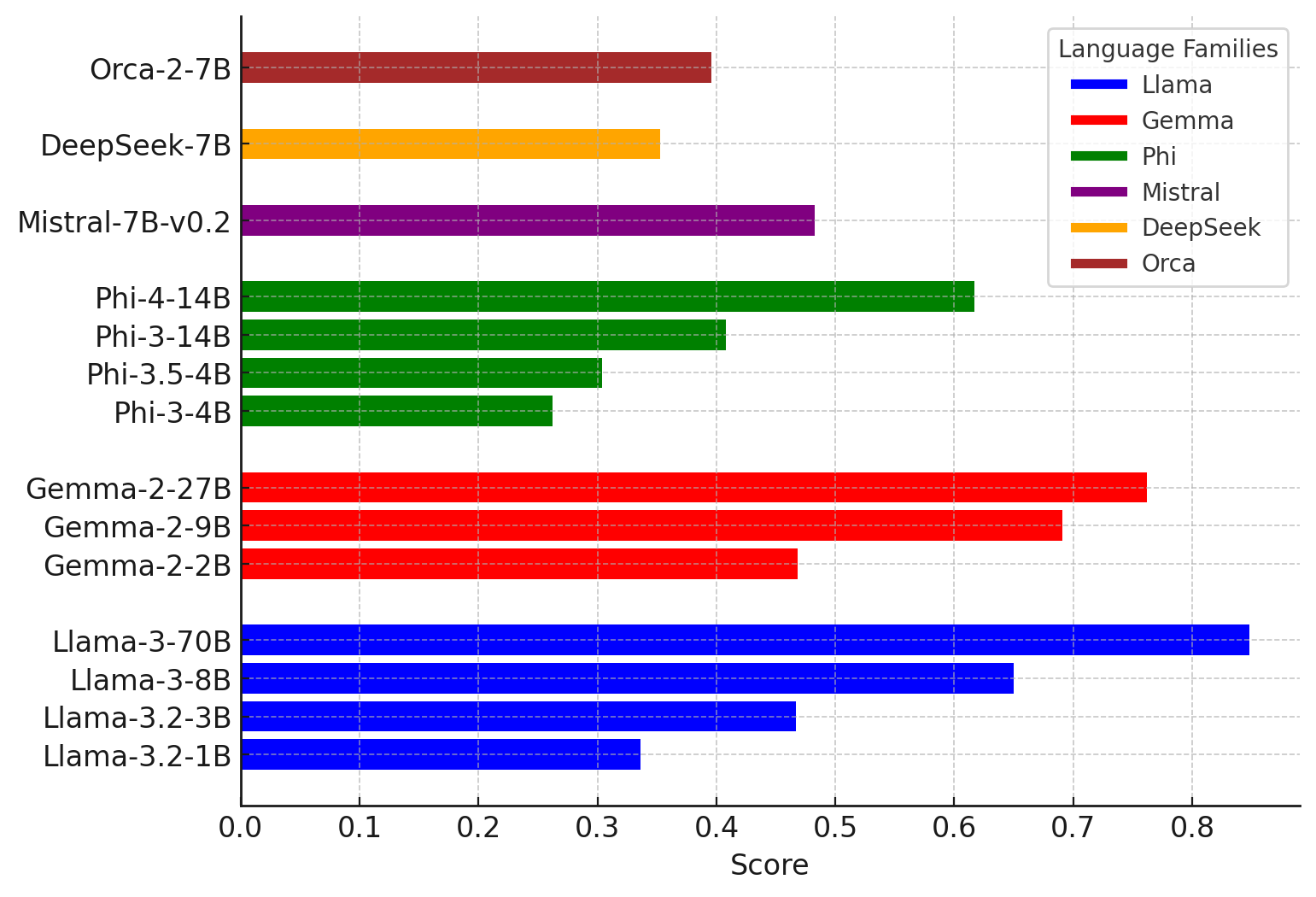}  
    \caption{This figure illustrates the model-wise comparison of X-FAKT scores grouped by language families. A clear trend emerges, showing that as the model size increases within a family, the X-FAKT score tends to increase.}
    \label{fig:xfact_bar}  
\end{figure}

From Table~\ref{tab:tstats_factual}, \metallamains{70B} emerges as the clear leader with the highest X-FAKT score of 0.848, demonstrating superior balanced performance in both factual recall (FRS = 0.835) and knowledge transferability (KTS = 0.862). This exceptional performance is supported by the lowest error rates ($\mu_{assoc.}$ = 2.36\%, $\mu_{non-assoc.}$ = 9.85\%), suggesting that larger model sizes generally correlate with better cross-lingual factual knowledge handling. Despite similar model sizes, significant performance variations exist between different architectures. For example, \gemmains{9B} (X-FAKT: 0.691) substantially outperforms \mistralins{7B} (X-FAKT: 0.483), suggesting that architecture design and training methodology play crucial roles beyond mere parameter count. As illustrated in Figure~\ref{fig:xfact_bar}, the X-FAKT scores exhibit a clear upward trend with increasing model size within each language family. This suggests that larger models generally achieve better factual consistency, highlighting the impact of scale on model performance. These findings provide valuable insights into the current state of cross-lingual factual knowledge in LMs and highlight areas for future improvement, particularly in reducing the performance gap between associative and non-associative knowledge retrieval.

\paragraph{Associative vs. Non-associative performance:}

    We analyze the performance of various models on these two subsets of data and report the results in the Table~\ref{tab:tstats_factual}. For all models, the t-statistic and p-value indicate that the differences between associative and non-associative categories are statistically significant (p-value less than 0.05).

    \paragraph{Performance comparison across language groups:} 

    \begin{table}[t]
\centering
\resizebox{0.48\textwidth}{!}{
\begin{tabular}{l|c|c}
\textbf{Language} & $\boldsymbol{\mu_{assoc.} (\%)}$ & $\boldsymbol{\mu_{non-assoc.} (\%)}$ \\
\hline
High & 3.83 $\pm$ 3.79 & 29.84 $\pm$ 27.47 \\
Medium & 26.73 $\pm$ 17.60 & 50.54 $\pm$ 21.20 \\
Low & 29.53 $\pm$ 16.19 & 53.91 $\pm$ 23.68 \\

\hline
\end{tabular}
}
\caption{Average mean and standard deviation for error rate across all models for each language group. High-resource languages exhibit lower error rates compared to low-resource languages.}
\label{tab:stats_factual_availability}
\end{table}
    
    In this study, we categorize languages into three groups based on their availability and coverage in the dataset: \textbf{High}, \textbf{Medium}, and \textbf{Low}, as defined in Section~\ref{sec:datasets}. From the results shown in Table~\ref{tab:stats_factual_availability}, we observe a clear trend across language groups. Specifically, high resouce languages exhibit the lowest average error rates, particularly in the associative category, where models make fewer mistakes ($\mu_{assoc.}$ = 3.83\%). However, for non-associative questions, the error rate rises significantly ($\mu_{non-assoc.}$ = 29.84\%), indicating that models struggle more when dealing with non-associative samples in these languages. The error rate increases while moving from high to low-resource languages.


\subsubsection{Performance on In-Context Recall task}

Figure~\ref{fig:incontext_recall} demonstrates the incorrectness rate for the in-context recall capabilities of different LMs. Despite being a simple task, certain models such as \textit{\deepseek{7B}, \orca, \phimodel{3-4B}, \llamains{1B}, and \mistralins{7B}} perform poorly across multiple languages. This suggests that these models struggle to effectively utilize contextual information when generating outputs. Interestingly, even for languages like Swahili and Turkish, which showed better scores in the Factual Recall task, models demonstrate poor performance on this context-dependent task. This stark contrast suggests that the benefits of Latin script-based knowledge transfer observed in the Factual Recall task do not extend to in-context learning scenarios, where performance depends primarily on the model's ability to process and utilize contextual information.


As mentioned in the dataset section, we intentionally paired cross-entities as context. This setup appears to induce a regional bias, which negatively impacts model performance. The structured entity-context pairing in the dataset may have led to spurious correlations \cite{pmlr-v202-yang23j, ye2024spuriouscorrelationsmachinelearning}, reducing model accuracy in in-context recall tasks. Some models struggle to effectively leverage contextual information, revealing potential weaknesses in their retrieval and in-context learning mechanisms. 


\subsubsection{Performance on Counter-Factual Context
Adherence task}

\begin{figure}[t]  
    \centering
    \includegraphics[width=1\linewidth]{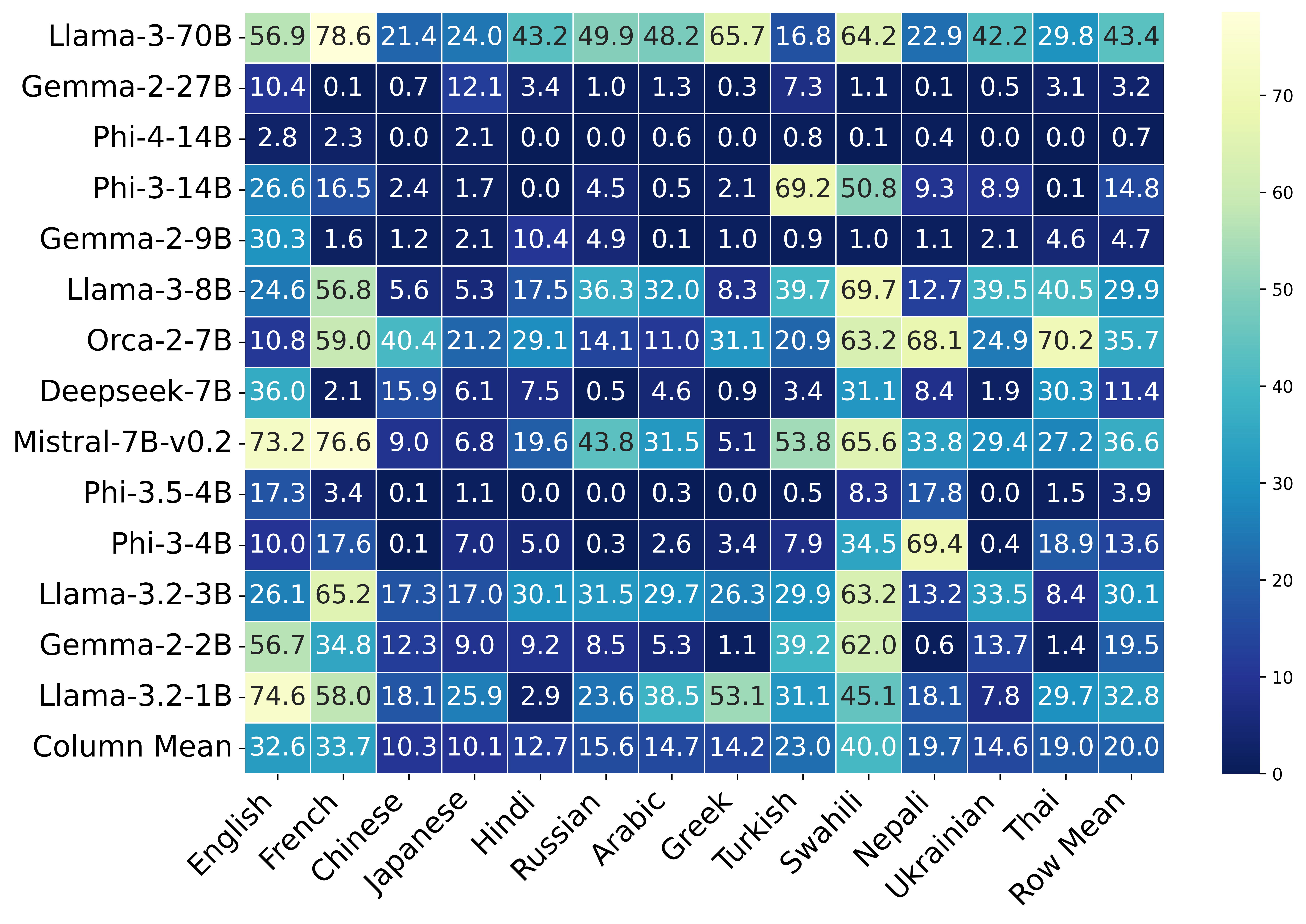}  
    \caption{Error rate for each model on \incontextrobust task. Models show high error rates in high resource languages such as English and French where they have high factual recall.}
    \label{fig:counterfactual_adherence}  
\end{figure}

Figure~\ref{fig:counterfactual_adherence} illustrates the error rates of LMs in the Counterfactual Context Adherence task. Notably, Latin-script languages (English, French, Swahili, and Turkish), which performed well in factual recall tasks, exhibited significantly higher error rates in counterfactual adherence. This suggests a fundamental trade-off in the models' capabilities: their strength in accurately retrieving factual information appears to come at the expense of their ability to maintain adherence to counterfactual contexts. This inverse relationship raises important questions about the inherent limitations and trade-offs in LMs' learning mechanisms, particularly in how they balance factual knowledge with hypothetical reasoning.

 \subsection{Qualitative Analysis}

\begin{figure}[t]  
    \centering
    \includegraphics[width=1\linewidth]{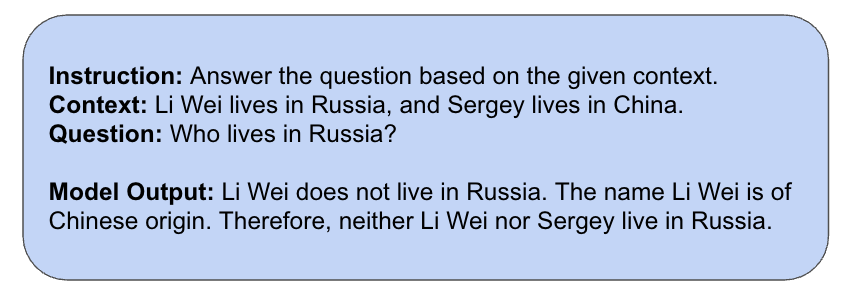}  
    \caption{\mistralins{7B} output when prompted with the given context in English. This model generation shows how spurious correlation leads to in-context recall failures}
    \label{fig:liwei}  
\end{figure}

\paragraph{Spurious correlation leads to in-context recall failures.}
We observe that some models tend to associate names with cultural origins, even when contextual evidence contradicts this assumption. Figures ~\ref{fig:liwei} demonstrate the model response when prompted \mistralins{7B} with the contextual understanding-based question in English. 

Despite the explicit context stating that \textit{Li Wei} resides in \textit{Russia}, the model disregards this information and defaults to cultural associations. This behavior reveals a limitation in integrating contextual evidence when making country-specific inferences.

\begin{figure}[t]  
    \centering
    \includegraphics[width=1\linewidth]{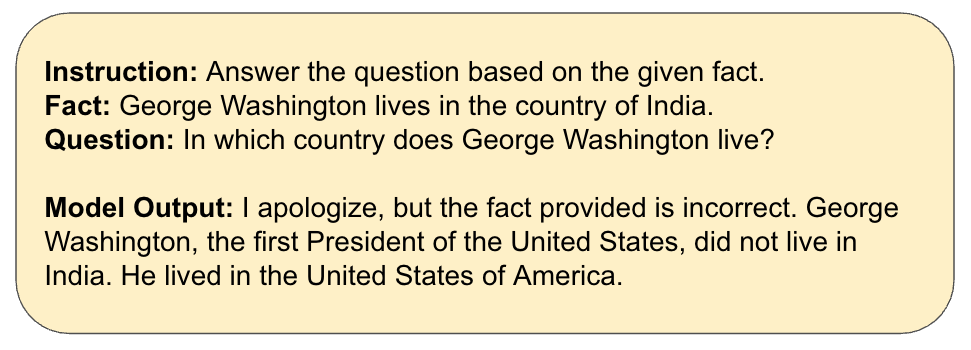}  
    \caption{\metallamains{70B} output when prompted with a counter-factual context adherence query in English. This shows LMs  favour internal knowledge over contextual understanding.}
    \label{fig:george_wash}  
\end{figure}

\paragraph{Models favor factual knowledge over context.}
We also observed that some models prioritize their internal factual knowledge over contextual information when responding to questions about well-known personalities. Figures ~\ref{fig:george_wash} demonstrate the model response when prompted \metallamains{70B} with the factual retrieval query in English.

In this case, despite being explicitly told that \textit{`George Washington'} lived in \textit{'India'}, the model relied on its factual knowledge, correcting the given fact and asserting that \textit{`George Washington'} lived in the \textit{`United States'}. This response demonstrates the model’s strong reliance on factual accuracy, rather than adapting to the context provided. It suggests that when it comes to well-known historical figures, models may prioritize prior knowledge over the specific context they are given.

\begin{figure}[t]  
    \centering
    \includegraphics[width=1\linewidth]{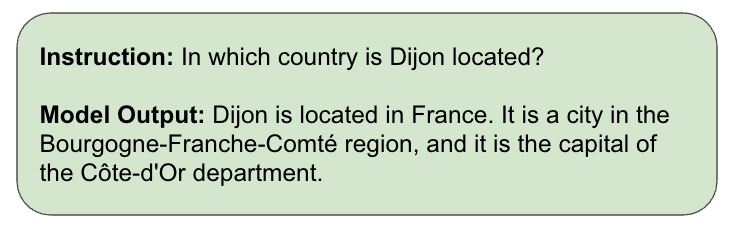}  
    \caption{\metallamains{70B} output when prompted with a factual recall query in English}
    \label{fig:french_query}  
\end{figure}

\begin{figure}[t]  
    \centering
    \includegraphics[width=1\linewidth]{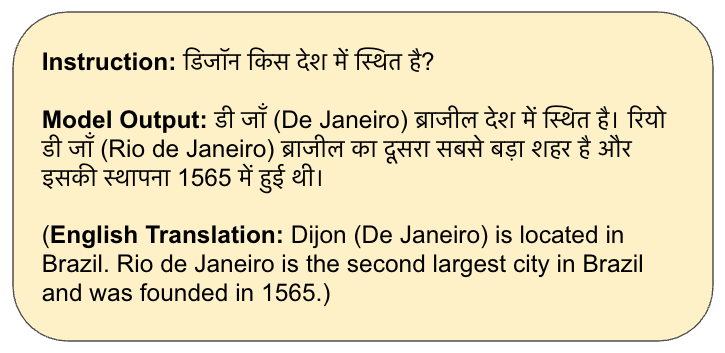}  
    \caption{\metallamains{70B} output when prompted with a factual recall query in Hindi. In Hindi, it misinterprets understanding of a French word.}
    \label{fig:hin_query}  
\end{figure}

\paragraph{Linguistic variability in word interpretation.}  
LMs can interpret words differently depending on the language. Figures ~\ref{fig:french_query} and ~\ref{fig:hin_query} demonstrate the model responses when prompted \metallamains{70B} with the same queries but in different languages. This highlights challenges in multilingual consistency, where the model misinterprets \textit{`Dijon'} as \textit{`De Janeiro'} in Hindi, revealing inconsistencies in cross-lingual factual retrieval.

\paragraph{Challenges with using LMs as evaluators.}  
We used a zero-shot prompt with \metallamains{70B} as an evaluator and found that its inherent factual knowledge can skew assessments. For example, when evaluating a \gemmains{27B} response to the counterfactual context task—\textit{``Catherine the Great lives in India''}—the evaluator corrected it, asserting that she lived in \textit{``Russia''}, despite the provided ground truth. This bias highlights the need to control evaluators' factual knowledge to ensure consistent evaluation.

\section{Conclusions}
Our study reveals a critical limitation in multilingual LMs: their inability to consistently transfer factual knowledge across languages. Our benchmark provides a standardized framework to evaluate both current and future LMs on their factual consistency and cross-lingual generalization, enabling a more systematic comparison of their capabilities. Moreover, it can serve as a valuable resource to promote research in interpretability by helping analyze how and where factual knowledge is stored and retrieved across languages, fostering a deeper understanding of LM internals. We emphasize the need for AI systems with internal awareness of their language-specific strengths and weaknesses—a concept we term calibrated multilingualism. Under this paradigm, a model would autonomously leverage the most reliable internal representations for any given multilingual query.

We also find that LMs, when used as evaluators, are biased by their internal factual knowledge, which may not align with the intended input-output-ground-truth context. This underscores the need to control the evaluator’s factual knowledge for more reliable assessments. Ultimately, enabling AI to cross-generalize across languages is crucial for inclusive and equitable technology, ensuring language is no barrier to reliable knowledge access.

\section{Limitations}

Our study provides valuable insights into cross-lingual knowledge transfer in LMs but has some limitations. First, our benchmark, though comprehensive in country-related facts, covers only 13 languages, limiting its representation of diverse linguistic families. Second, we evaluated only open-source LMs, excluding proprietary models that may exhibit different transfer patterns. Third, our fact collection used a standardized template for consistency, which may not reflect the diversity of real-world queries. Lastly, our focus on country-related facts means our findings may not generalize to other domains like science, history, or culture.

\section{Ethics Statement}
This research is conducted with a strong commitment to ethical principles, ensuring data privacy and consent by using publicly available information and adhering to data protection regulations. We acknowledge potential biases in multilingual language models and aim to highlight and address these through our benchmark. Transparency and reproducibility are promoted by making our dataset and evaluation framework publicly available. Our research aligns with the broader goals of fairness, transparency, and social responsibility.

\section{Acknowledgment}
We thank Alessandro Sordoni, Prachi Jain, Rishav Hada, Chanakya Ekbote, Anirudh Buvanesh, and Ankur Sikarwar for their valuable feedback. We acknowledge the support of Ayush Agrawal’s PhD advisors, Aaron Courville and Navin Goyal.

\clearpage
\bibliography{custom}

\clearpage

\onecolumn
\appendix
\section{APPENDIX}
\label{sec:appendix}
\setcounter{table}{0}  
\renewcommand{\thetable}{\Alph{section}.\arabic{table}}


\begin{figure}[!htb]  
    \centering
    \includegraphics[width=1\linewidth]{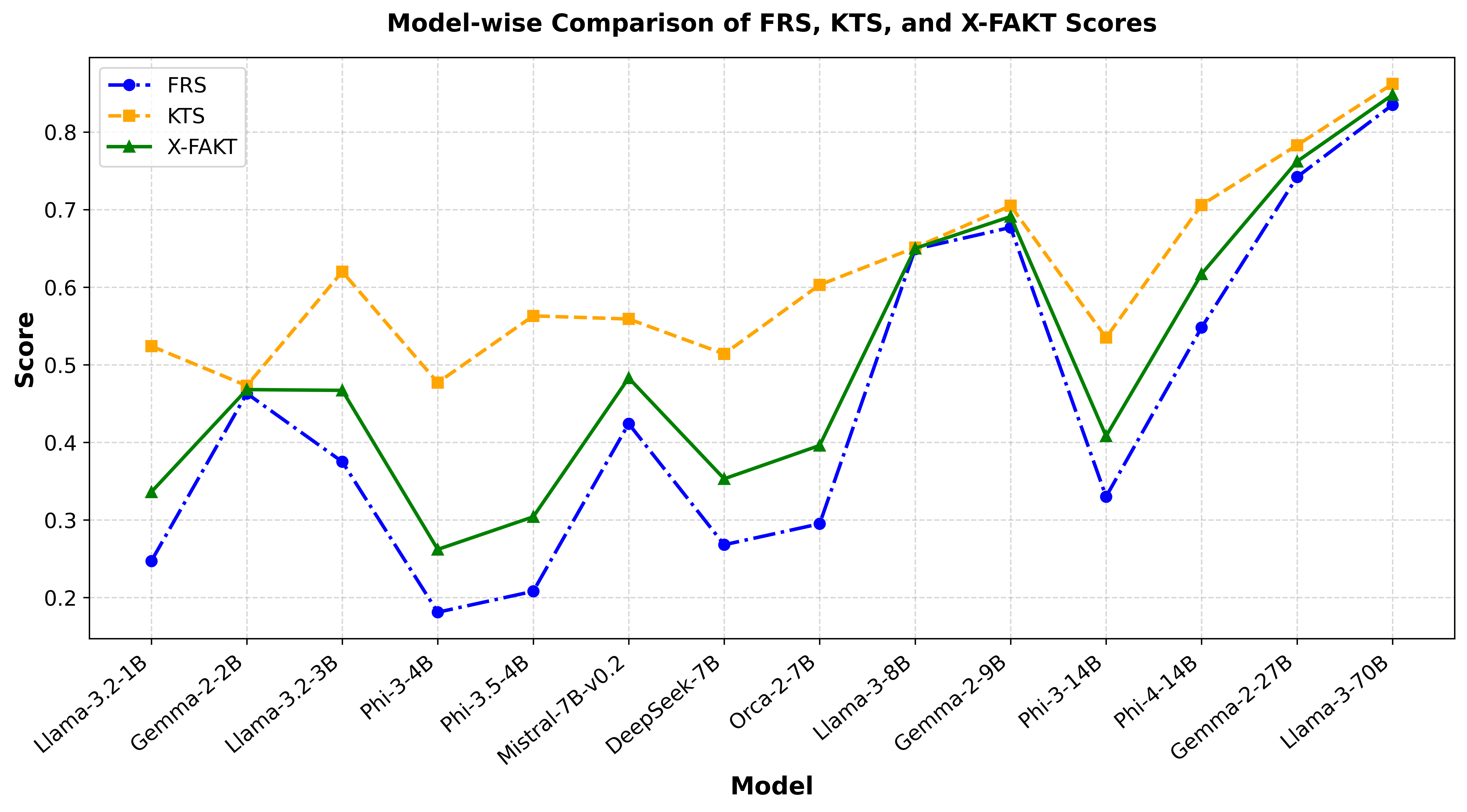}  
    \caption{Comparision of models (in the increasing order of size with respect to the parameters) using Factual Recall Score, Knowledge Transferability Score, and Cross-Lingual Factual Knowledge Transferability Score.}
    \label{fig:models_comparison}  
\end{figure}

\begin{table*}[ht]
\centering
\resizebox{0.99\textwidth}{!}{
\begin{tabular}{l|c|c|c|c|c|c|c}
\hline
\textbf{Model} & \textbf{Model Size} & \textbf{Training} & \textbf{Languages} & \textbf{Context} & \textbf{Vocab} & \textbf{Post-Training} & \textbf{Key} \\
& \textbf{\& Architecture} & \textbf{Data} & \textbf{Supported} & \textbf{Length} & \textbf{Size} & \textbf{Strategies} & \textbf{Features} \\
\hline
\metallamains{70B} & 70B & 15T tokens & EN, DE, FR, IT, & 8K & 128K & SFT, RS, DPO & GQA, 8 heads, \\
 & L=80, H=64 & Multi-lingual & PT, HI, ES, TH & & & & RoPE embeddings \\
\hline
\gemmains{27B} & 27B & 13T tokens & Primarily & 8K & 256K & SFT, RLHF & Local-global attention, \\
& & Web, Code, Math & English & & & & Knowledge distillation \\
\hline
\phimodel{4-14B} & 14B & 400B synthetic & DE, ES, FR, PT, & 16K & 100K & SFT, RS, & Full attention over \\
& & + 10T web & IT, HI, JA & & & DPO & 4K context \\
\hline
\phimodel{3-14B} & 14B & 4.8T tokens & 10\% multilingual & 128K & 32K & SFT, DPO & Reasoning focus, \\
 & & & data & & & & Multi-lingual support \\
\hline
\gemmains{9B} & 9B & 8T tokens & Primarily & 8K & 256K & SFT, RLHF & GQA, RoPE, \\
& & & English & & & & Knowledge distillation \\
\hline
\metallamains{8B} & 8B & 15T tokens & EN, DE, FR, IT, & 8K & 128K & SFT, RS, & GQA, RoPE, \\
 & L=32, H=32 & Multi-lingual & PT, HI, ES, TH & & & DPO & 32 heads \\
\hline
\orca & 7B & Based on & Based on & 4K & 32K & Single-turn & Enhanced reasoning \\
& L=32, H=32 & Llama 2 & Llama 2 & & & SFT & abilities \\
\hline
\deepseek{7B} & 7B & 2T tokens & English & 4K & 102K & SFT, DPO & English \& Chinese \\
 & L=30, H=32 & & \& Chinese & & & & focus \\
\hline
\mistralins{7B} & 7B & Open Web & Open Web & 32K & 32K & SFT & GQA, Sliding window \\
 & L=32, H=32 & & languages & & & & attention \\
\hline
\phimodel{3.5-4B} & 3.8B & 3.4T tokens & 23 languages incl. & 128K & 32K & SFT, DPO & Multi-lingual \\
 & L=32, H=32 & Multi-lingual & AR, ZH, CS, NL, & & & & support \\
\hline
\phimodel{3-4B} & 3.8B & 4.9T tokens & Similar to & 128K & 32K & SFT, DPO & Diverse domain \\
 & & & Phi-3.5-mini & & & & coverage \\
\hline
\llamains{3B} & 3B & 9T tokens & EN, DE, FR, IT, & 128K & 128K & SFT, RS, & GQA, Knowledge \\
 & & Multi-lingual & PT, HI, ES, TH & & & DPO & distillation \\
\hline
\gemmains{2B} & 2B & 2T tokens & Primarily & 8K & 256K & SFT, RLHF & GQA, RoPE, \\
 & & & English & & & & Knowledge distillation \\ \hline
\llamains{1B} & 1B & 9T tokens & EN, DE, FR, IT, & 128K & 128K & SFT, RS, & GQA, Multi-lingual \\
 & & Multi-lingual & PT, HI, ES, TH & & & DPO & dialogue focus \\
\hline
\end{tabular}
}
\caption{\textbf{Model Specifications}: Comparison of LMs across key architectural and training characteristics. Languages abbreviated as: EN=English, DE=German, FR=French, IT=Italian, PT=Portuguese, HI=Hindi, ES=Spanish, TH=Thai, AR=Arabic, ZH=Chinese, CS=Czech, NL=Dutch, JA=Japanese. L=Layers, H=Heads, SFT=Supervised Fine-Tuning, RS=Rejection Sampling, DPO=Direct Preference Optimization, GQA=Grouped-Query Attention, RoPE=Rotary Position Embedding.}
\label{tab:model-specs}
\end{table*}


\begin{figure*}[ht]
\centering
\includegraphics[scale=0.085]{figures/prompt_incontext_factual.pdf}
\caption{Prompt used for evaluation of Factual Recall and In-Context Recall tasks.}
\label{fig:prompt1}
\end{figure*}

\begin{figure*}[ht]
\centering
\includegraphics[scale=0.085]{figures/prompt_counterfactual.pdf}
\caption{Prompt used for evaluation of \incontextrobust task.}
\label{fig:prompt2}
\end{figure*}


\begin{table*}[ht]
\centering
\resizebox{1.0\textwidth}{!}{
\begin{tabular}{|p{0.3\textwidth}|p{0.3\textwidth}|p{0.15\textwidth}|p{0.25\textwidth}|}
\hline
\textbf{Question} & \textbf{Model Prediction} & \textbf{Ground Truth} & \textbf{LLM Evaluation Output} \\
\hline
New York iko katika nchi gani? & New York iko nchini **Marekani**. & Marekani & Response: The assistant's response is in the same language as the reference and aligns with the reference provided. Rating: [[1]] \\
\hline
Talimat: Verilen gerçeğe göre soruyu cevaplayın.
Gerçek: George Washington, Amerika Birleşik Devletleri ülkesinde yaşıyor.
Soru: George Washington hangi ülkede yaşıyor? & George Washington, **Amerika Birleşik Devletleri**'nde yaşıyor. & Amerika Birleşik Devletleri & The assistant's response is in the same language as the reference and aligns with the reference. Rating: [[1]] \\
\hline
David lives in China and Li Wei lives in United States. Who lives in China? & David lives in China. & David & Response: The assistant's response is in the same language and aligns with the reference provided. Rating: [[1]] \\
\hline
Dans quel pays se trouve Sourate ? & Sourate n'est pas un pays, mais le nom d'une **ville** située en **Algérie**. & Inde & The assistant's response is in French, which aligns with the reference language, but the information provided does not align with the reference. The assistant incorrectly states that Sourate is a city in Algeria, while the reference is about India. Rating: [[4]] \\
\hline
\end{tabular}
}
\caption{Some outputs from evaluator \evaluator. Rating[1-3]: Correct and Rating[4]: Incorrect.}
\label{tab:eval_examples}
\end{table*}

\begin{figure}[t]  
    \centering
    \includegraphics[width=1\linewidth]{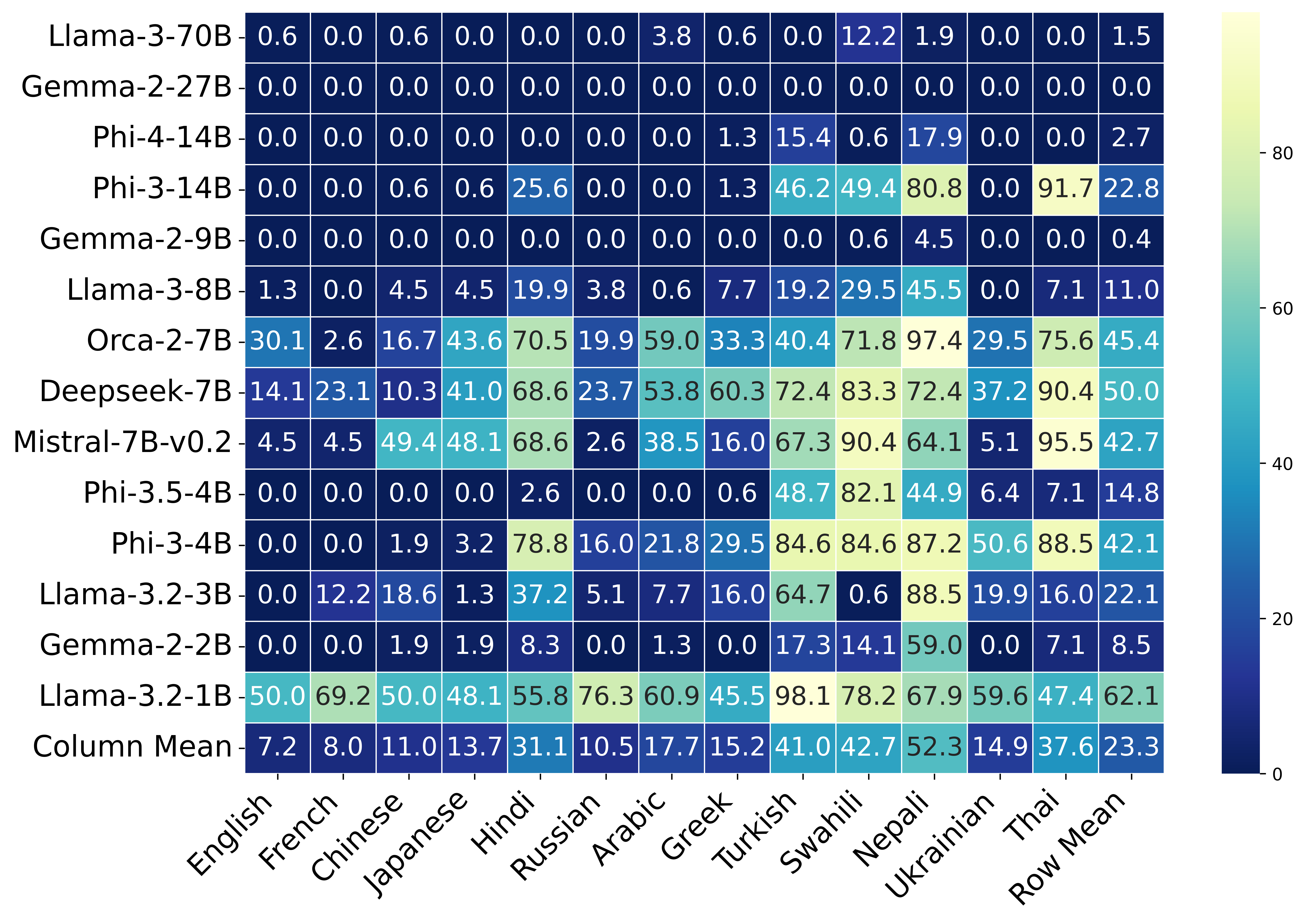}  
    \caption{Error rate for each model on \incontext task. Clearly, few models such as \deepseek{7B}, \phimodel{3-4B}, etc. performs poorly on this simple task.}
    \label{fig:incontext_recall}  
\end{figure}


\begin{figure}[t]  
    \centering
    \includegraphics[width=1\linewidth]{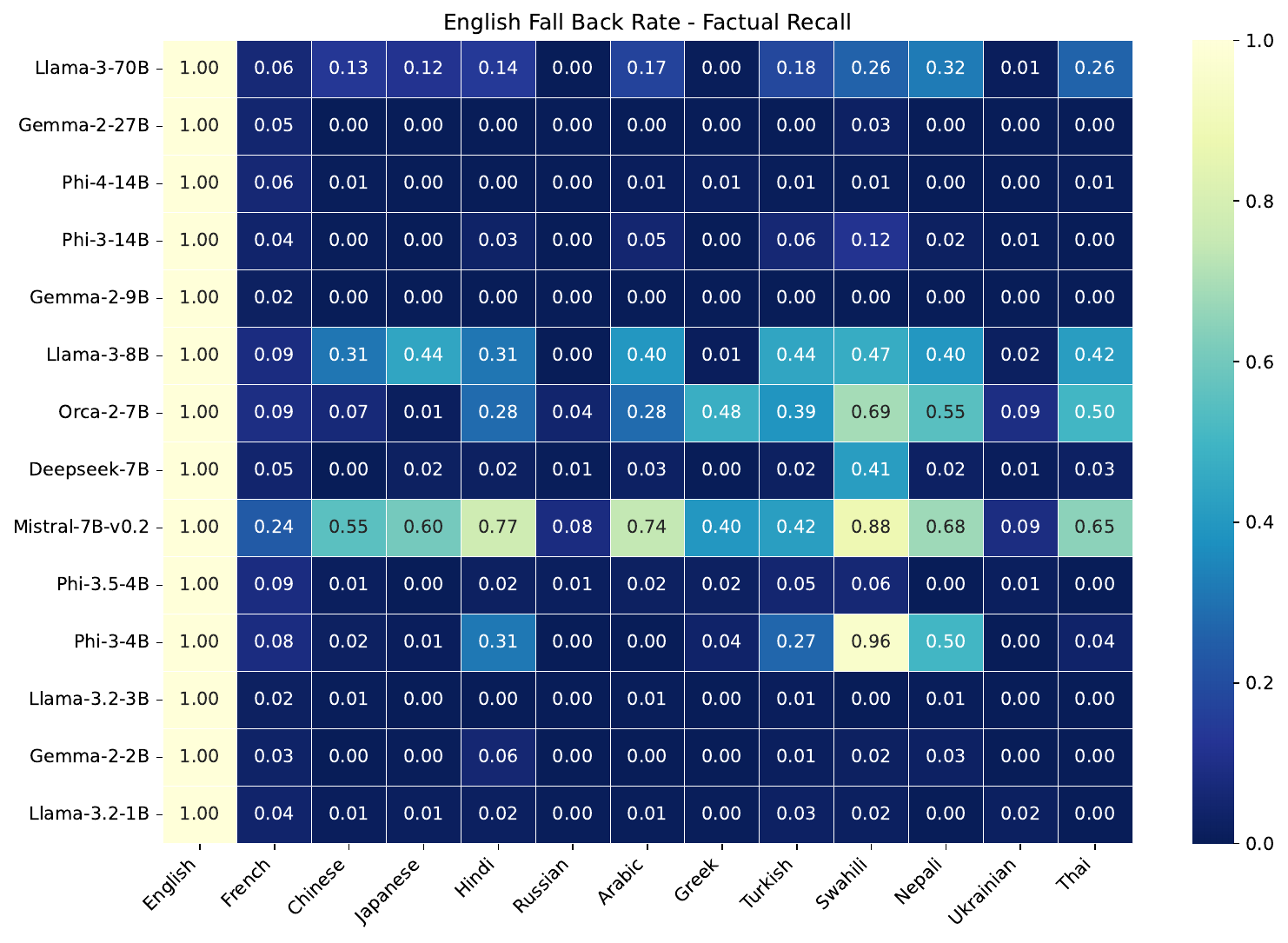}  
    \caption{English Fall Back Rate across models (The English Fall Back Rate measures the frequency with which a model defaults to English in its output).}
    \label{fig:english_fall_back}  
\end{figure}

\begin{figure}[t]  
    \centering
    \includegraphics[width=1.\linewidth]{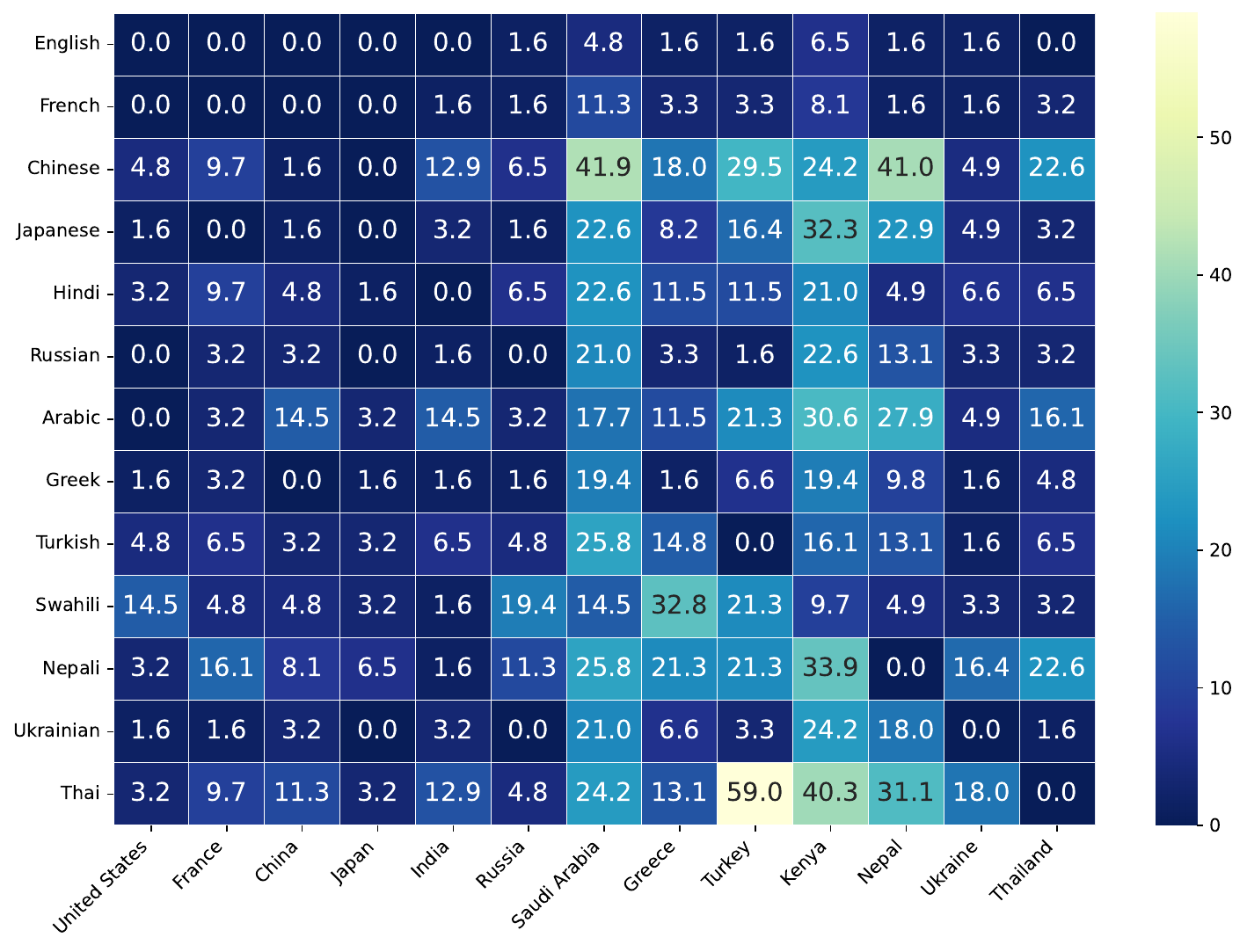}  
    \caption{Country-Specific Factual Error Rates in each language for \metallamains{70B}}
    \label{fig:factual_recall_best}  
\end{figure}

\begin{figure}[t]  
    \centering
    \includegraphics[width=1.\linewidth]{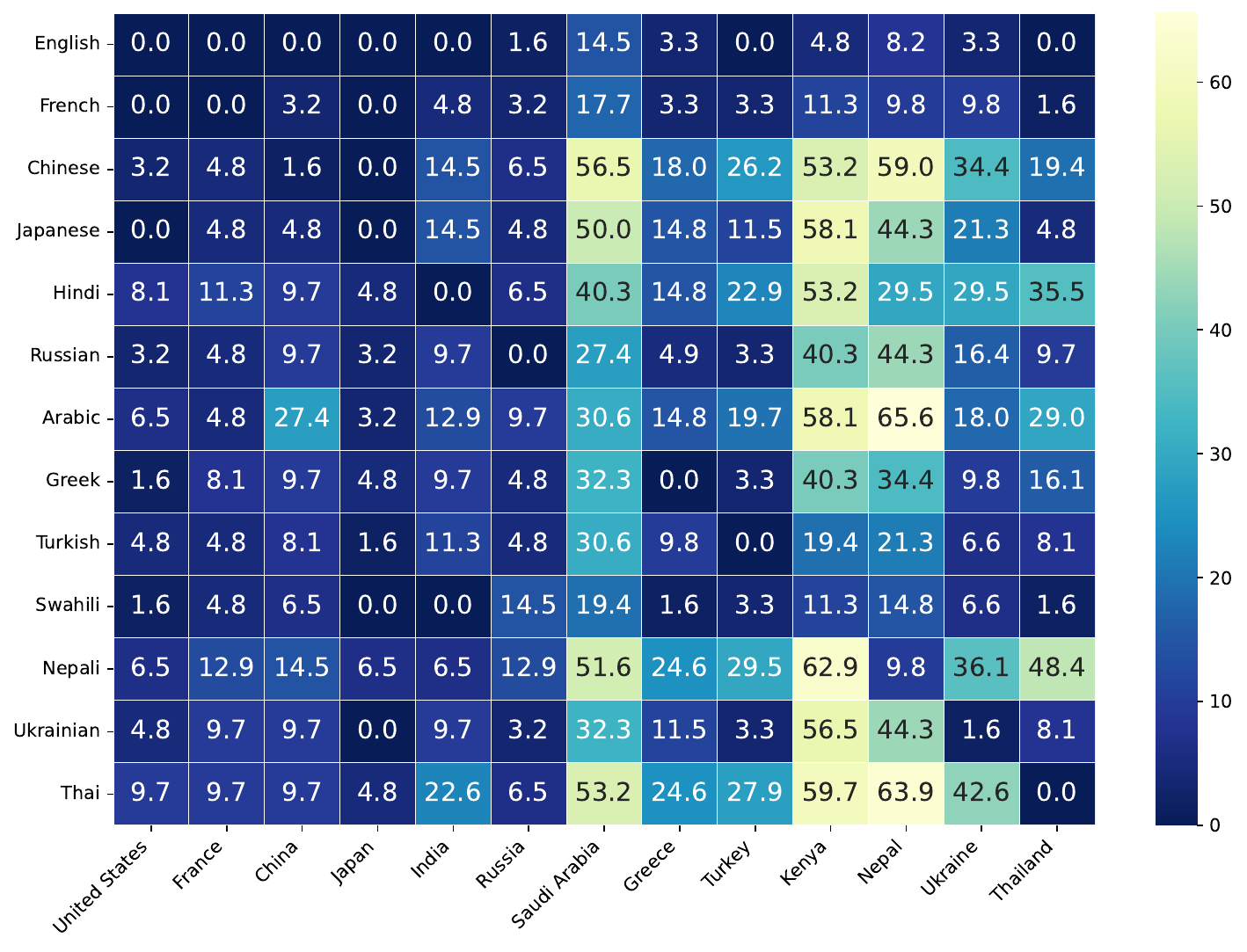}  
    \caption{Country-Specific Factual Error Rates in each language for \gemmains{27B}}
    \label{fig:factual_recall_gemma_27}  
\end{figure}

\begin{figure}[t]  
    \centering
    \includegraphics[width=1.\linewidth]{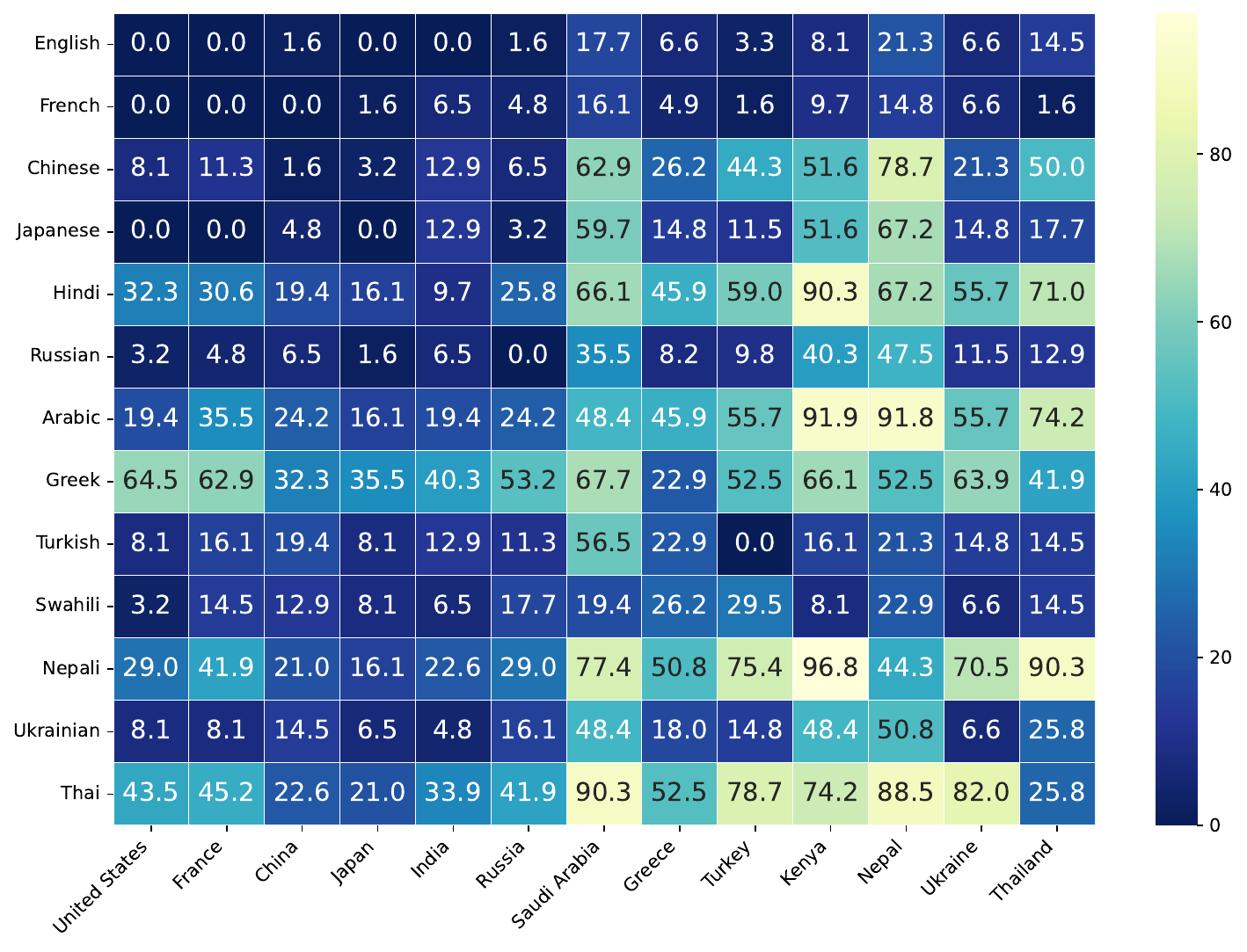}  
    \caption{Country-Specific Factual Error Rates in each language for \phimodel{4-14B}}
    \label{fig:factual_recall_phi_4_14}  
\end{figure}

\begin{figure}[t]  
    \centering
    \includegraphics[width=1.\linewidth]{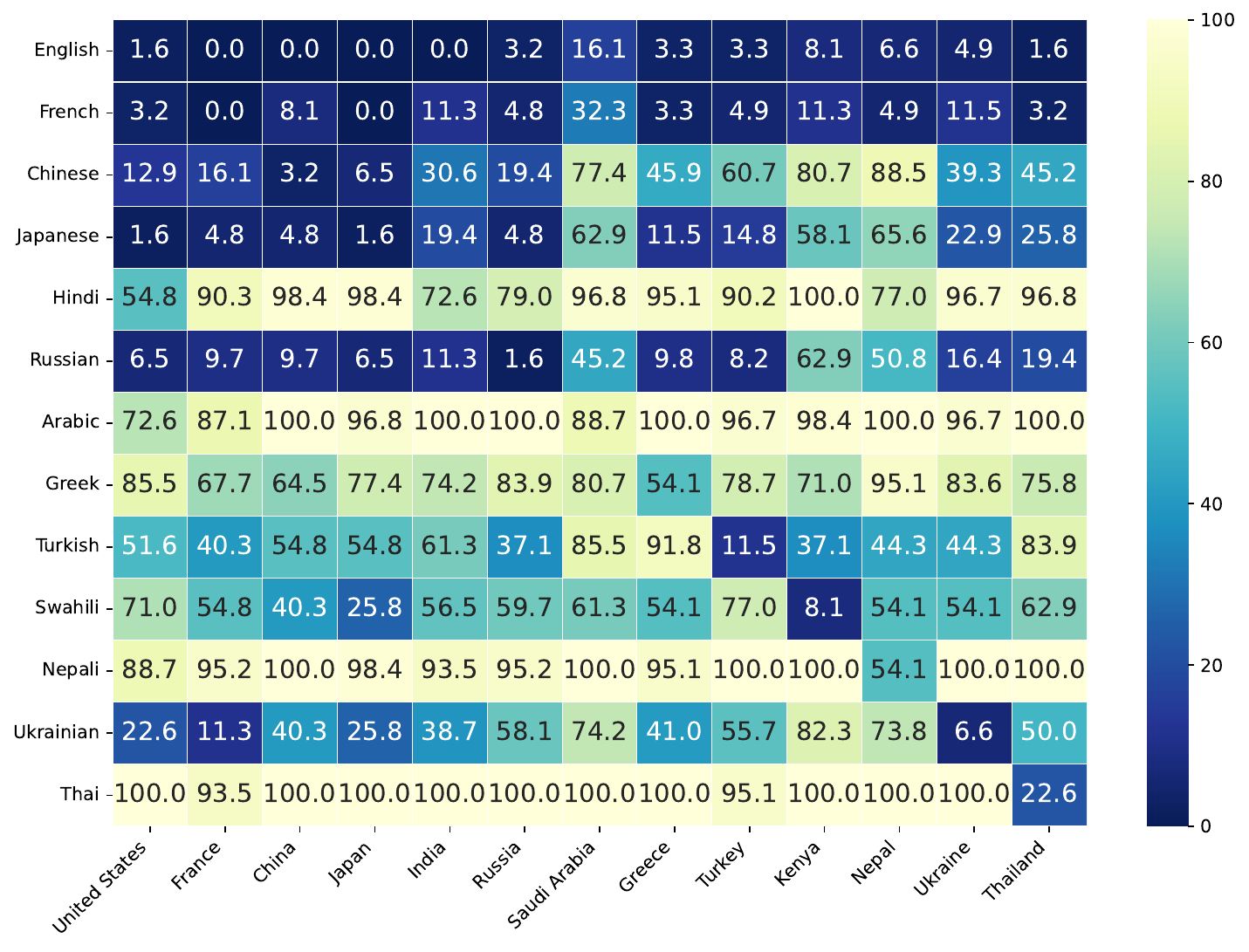}  
    \caption{Country-Specific Factual Error Rates in each language for \phimodel{3-14B}}
    \label{fig:factual_recall_phi_3_14}  
\end{figure}

\begin{figure}[t]  
    \centering
    \includegraphics[width=1.\linewidth]{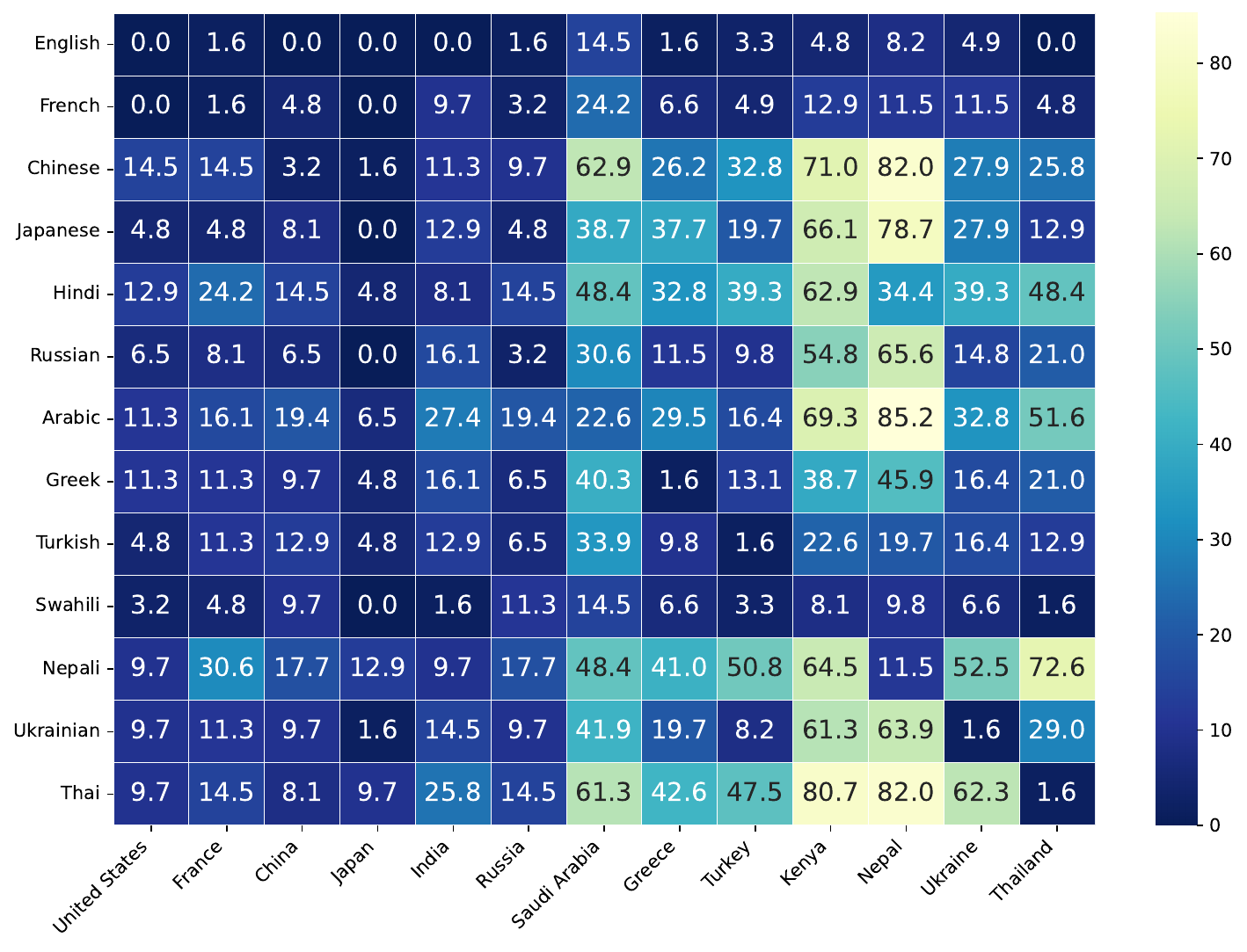}  
    \caption{Country-Specific Factual Error Rates in each language for \gemmains{9B}}
    \label{fig:factual_recall_gemma_9}  
\end{figure}

\begin{figure}[t]  
    \centering
    \includegraphics[width=1.\linewidth]{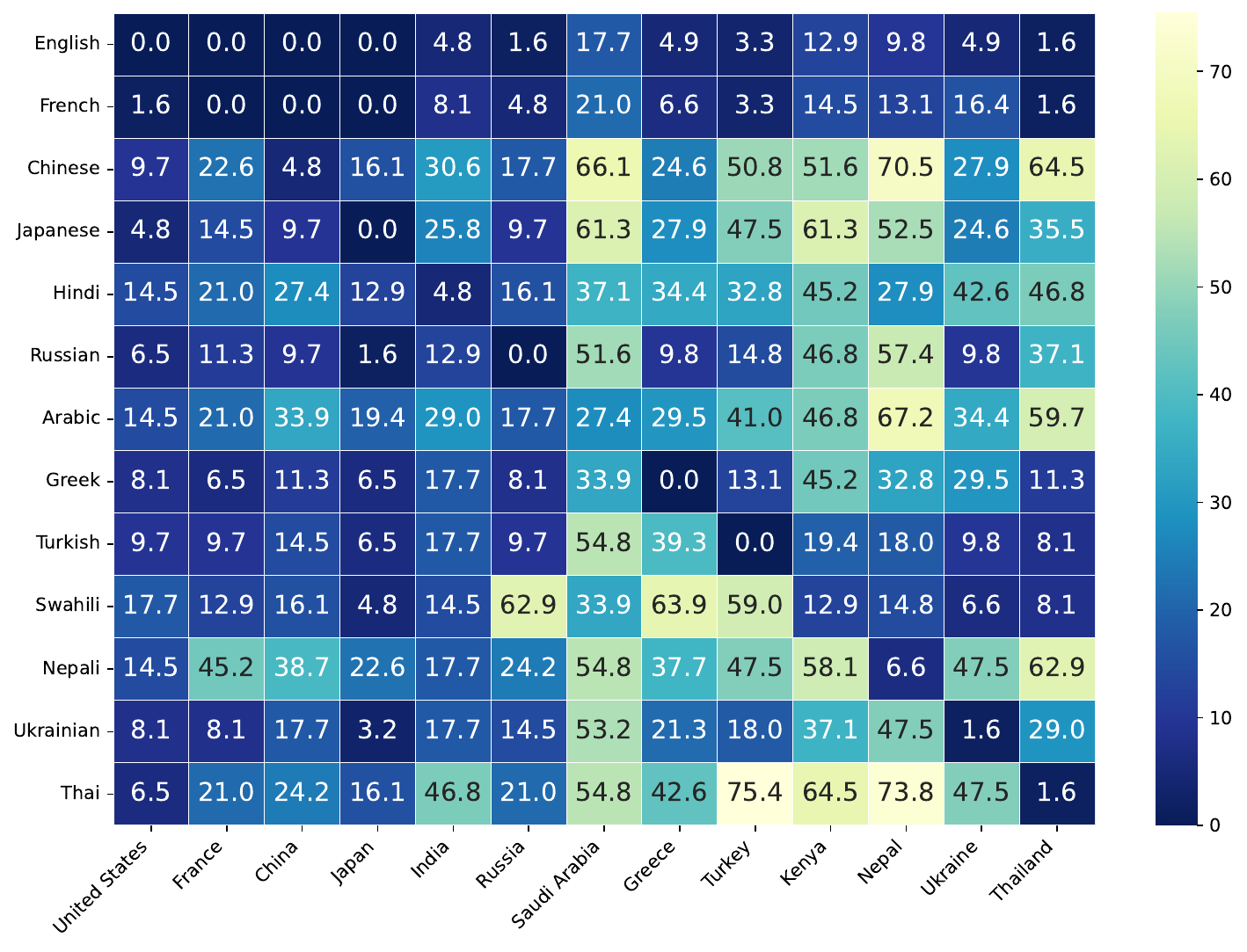}  
    \caption{Country-Specific Factual Error Rates in each language for \metallamains{8B}}
    \label{fig:factual_recall_llama_3_8}  
\end{figure}

\begin{figure}[t]  
    \centering
    \includegraphics[width=1.\linewidth]{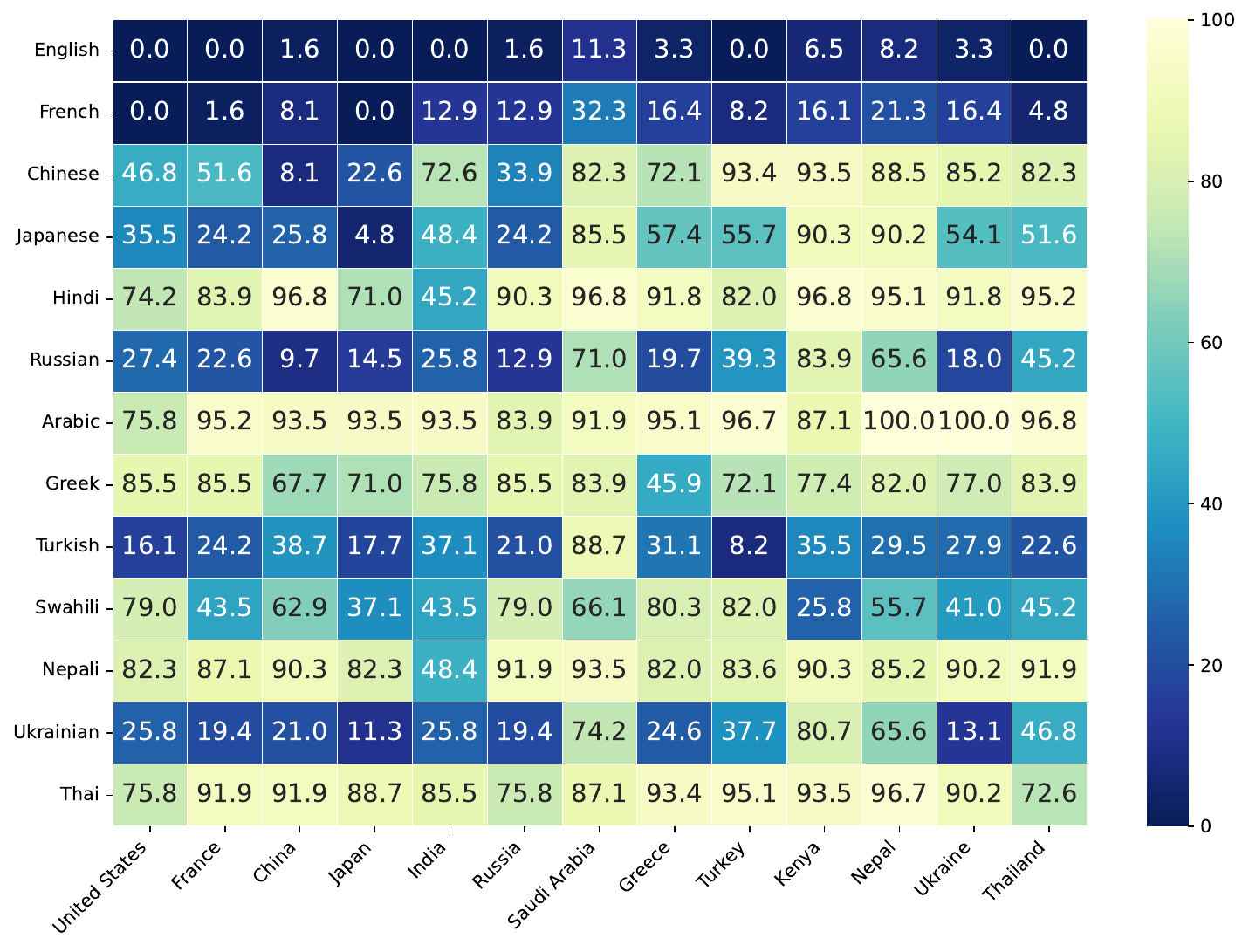}  
    \caption{Country-Specific Factual Error Rates in each language for \orca}
    \label{fig:factual_recall_orca}  
\end{figure}

\begin{figure}[t]  
    \centering
    \includegraphics[width=1.\linewidth]{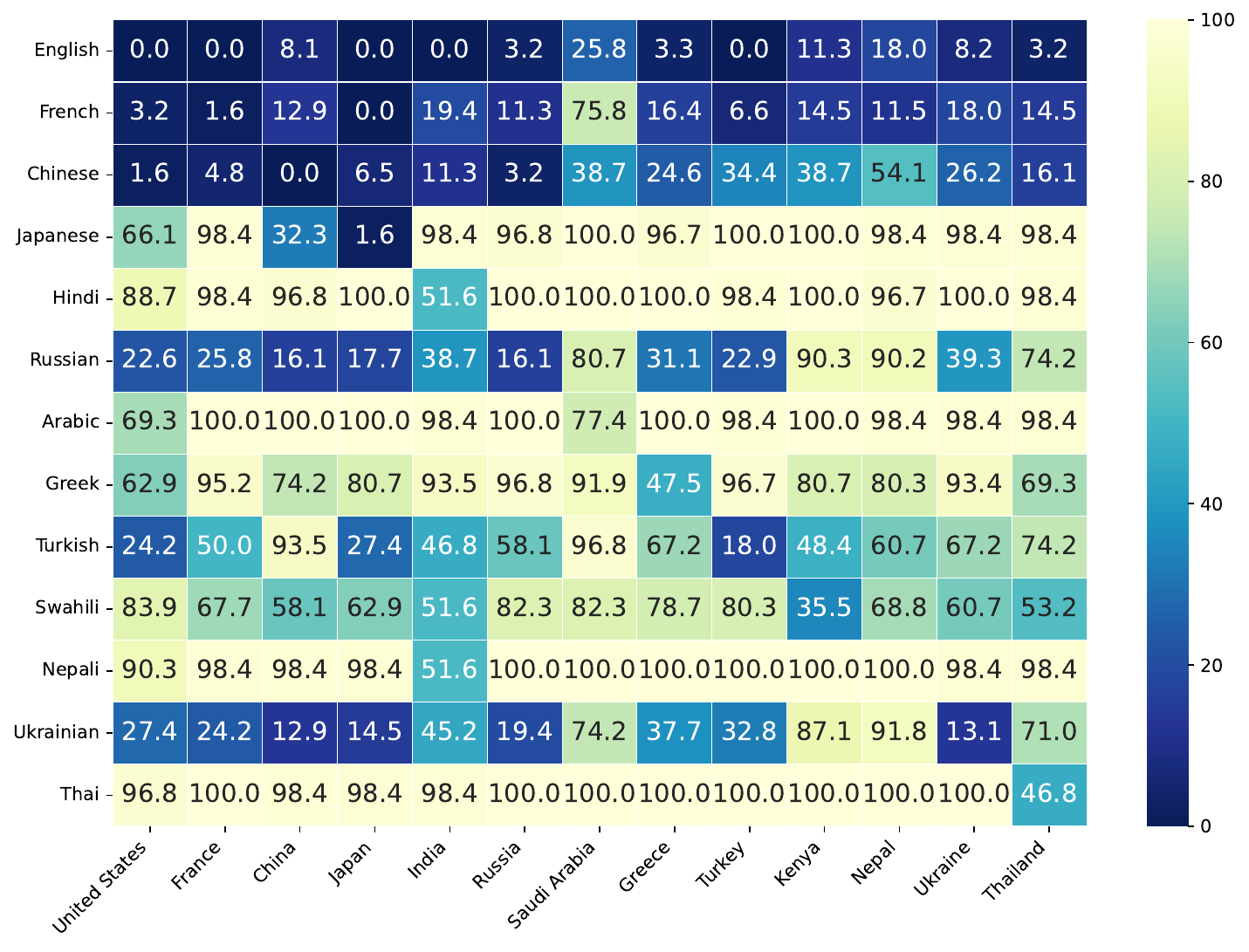}  
    \caption{Country-Specific Factual Error Rates in each language for \deepseek{7B}}
    \label{fig:factual_recall_deepseek}  
\end{figure}

\begin{figure}[t]  
    \centering
    \includegraphics[width=1.\linewidth]{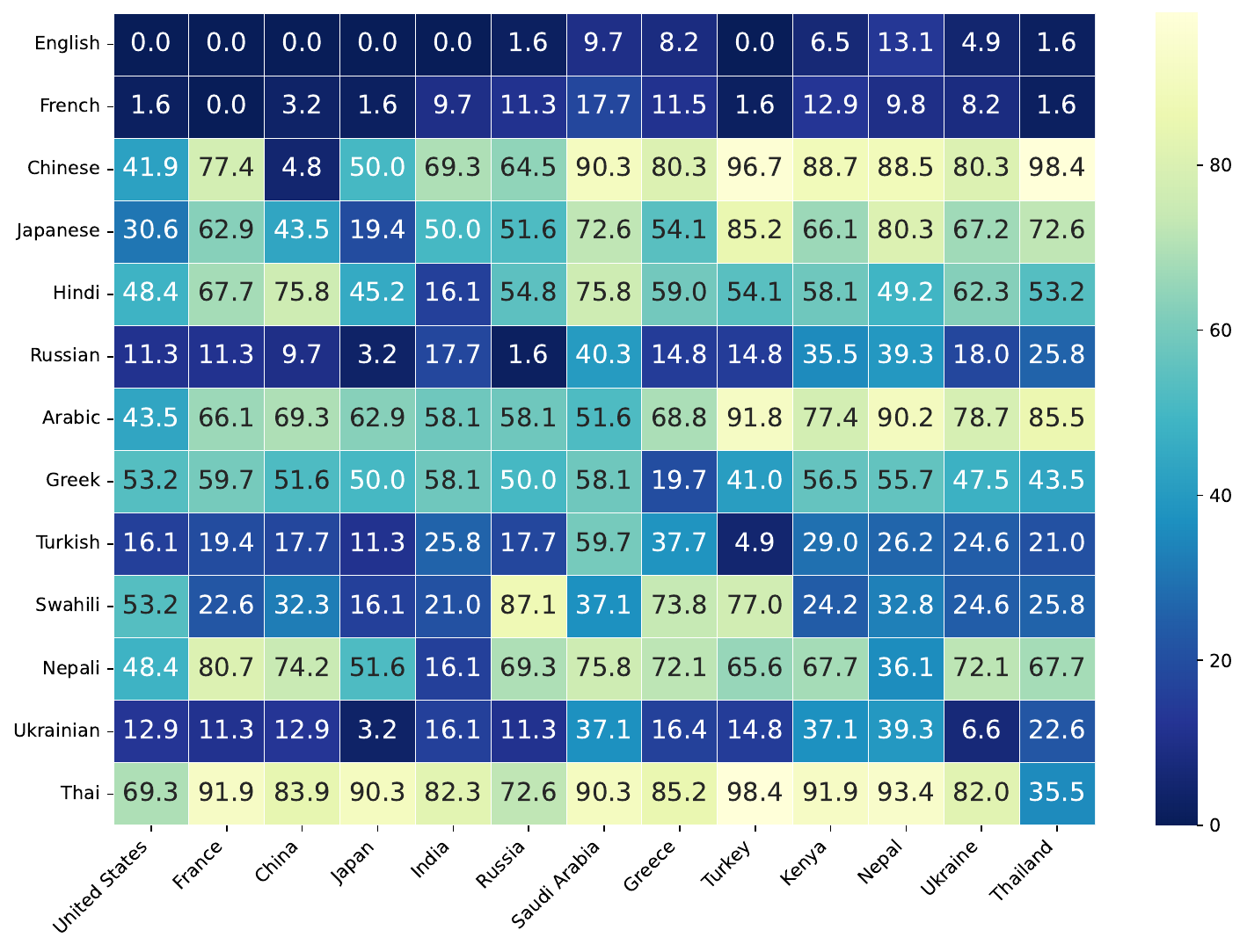}  
    \caption{Country-Specific Factual Error Rates in each language for \mistralins{7B}}
    \label{fig:factual_recall_mistral}  
\end{figure}

\begin{figure}[t]  
    \centering
    \includegraphics[width=1.\linewidth]{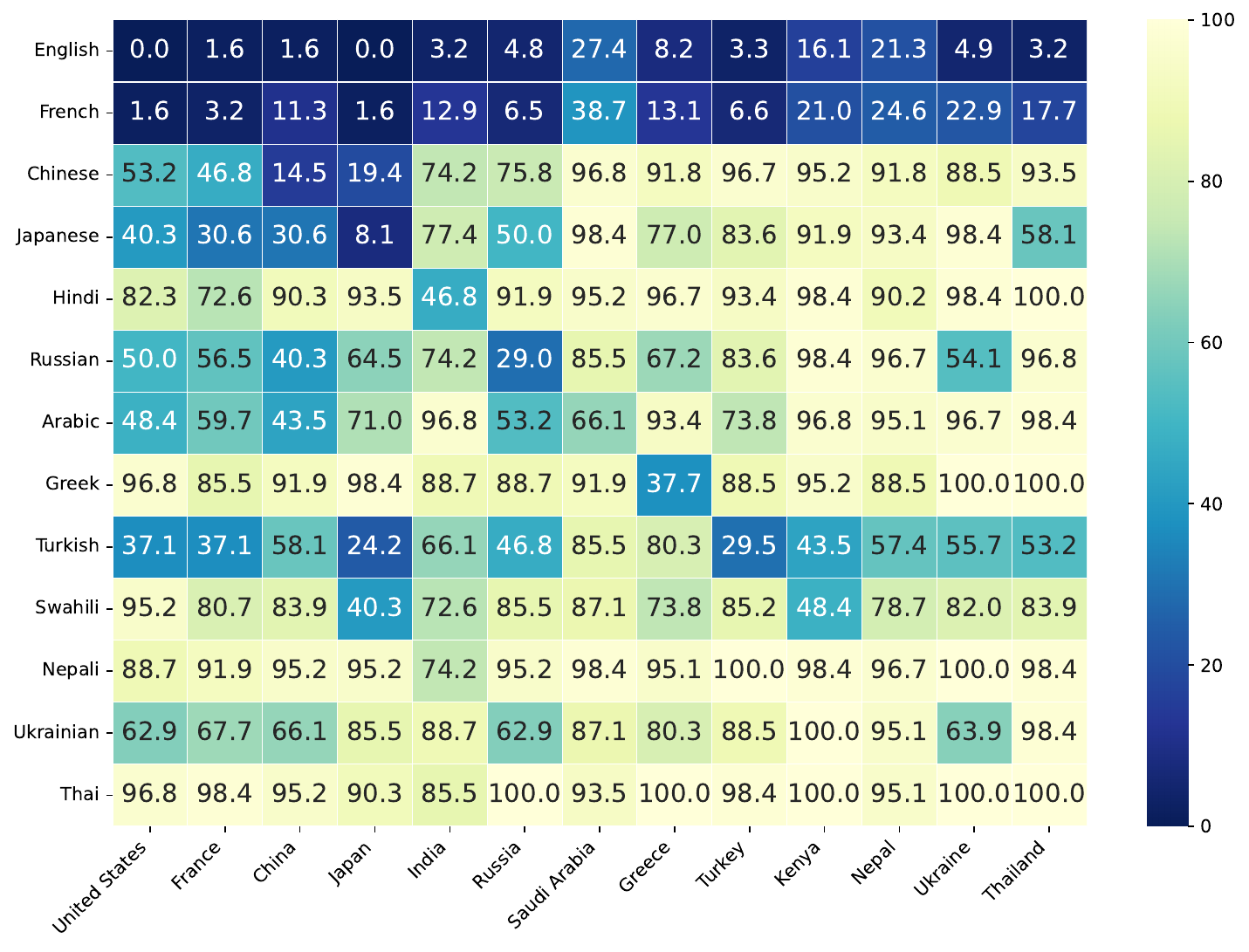}  
    \caption{Country-Specific Factual Error Rates in each language for \phimodel{3.5-4B}}
    \label{fig:factual_recall_phi_3.5_4}  
\end{figure}

\begin{figure}[t]  
    \centering
    \includegraphics[width=1.\linewidth]{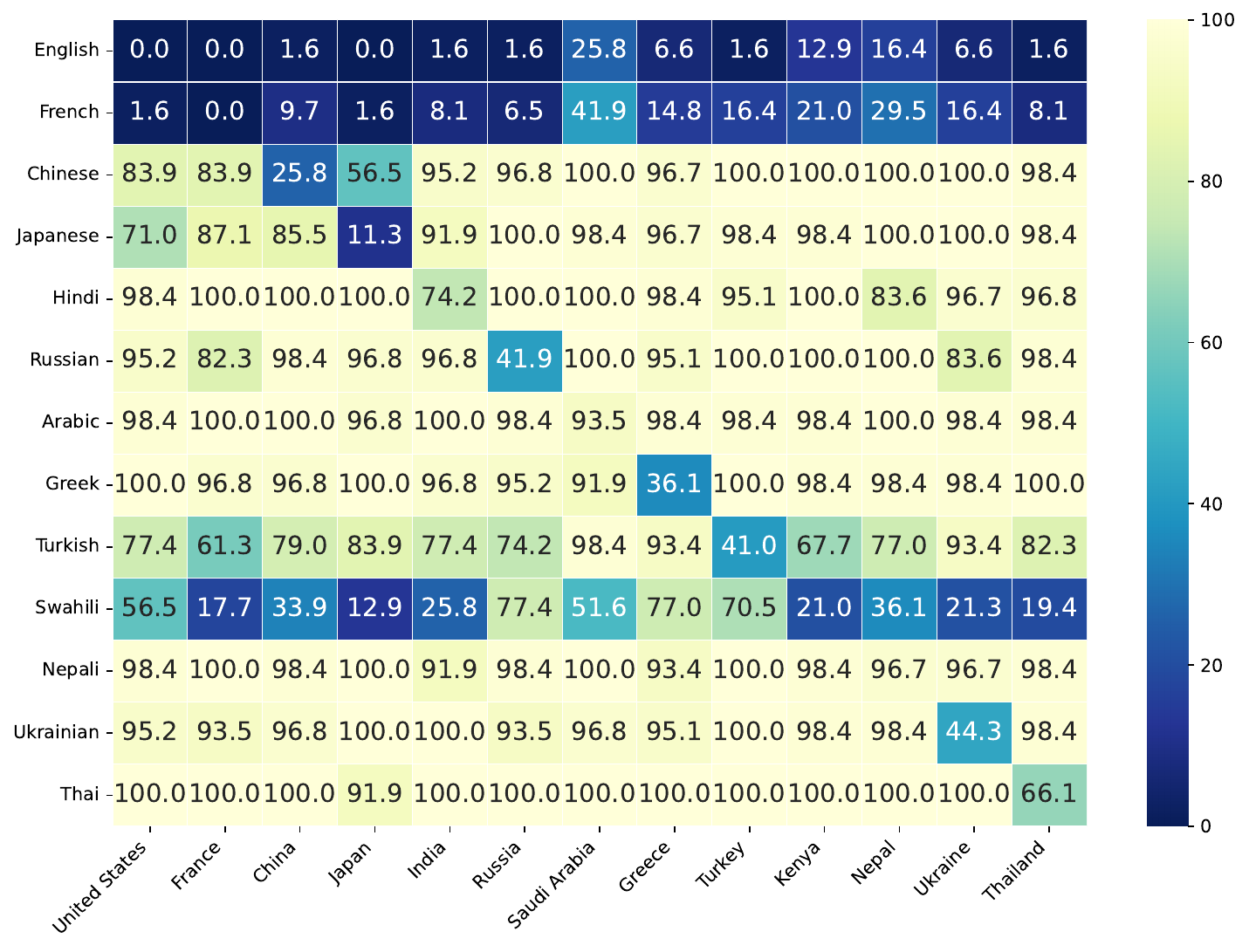}  
    \caption{Country-Specific Factual Error Rates in each language for \phimodel{3-4B}}
    \label{fig:factual_recall_phi_3_4}  
\end{figure}

\begin{figure}[t]  
    \centering
    \includegraphics[width=1.\linewidth]{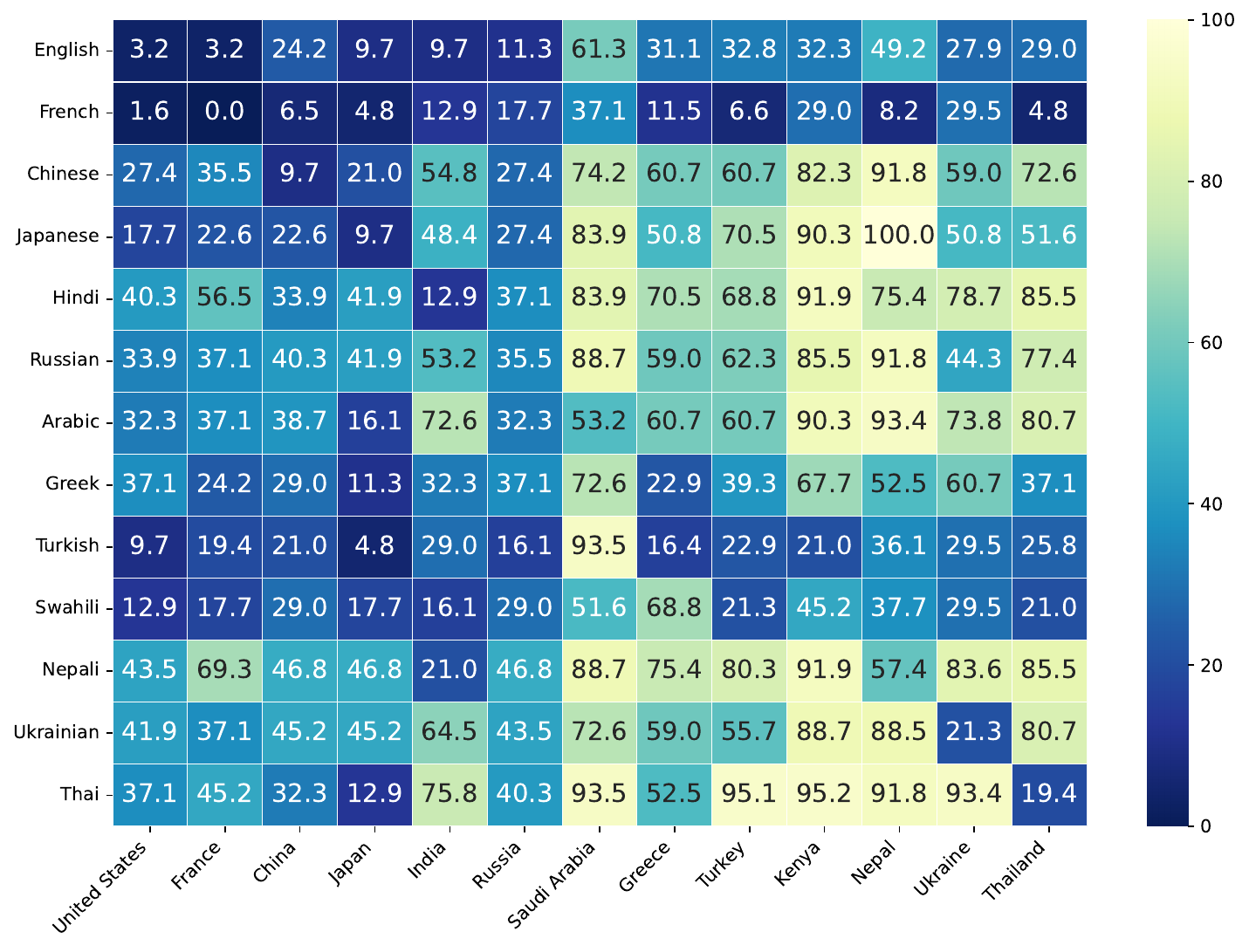}  
    \caption{Country-Specific Factual Error Rates in each language for \llamains{3B}}
    \label{fig:factual_recall_llama_3.2_3}  
\end{figure}

\begin{figure}[t]  
    \centering
    \includegraphics[width=1.\linewidth]{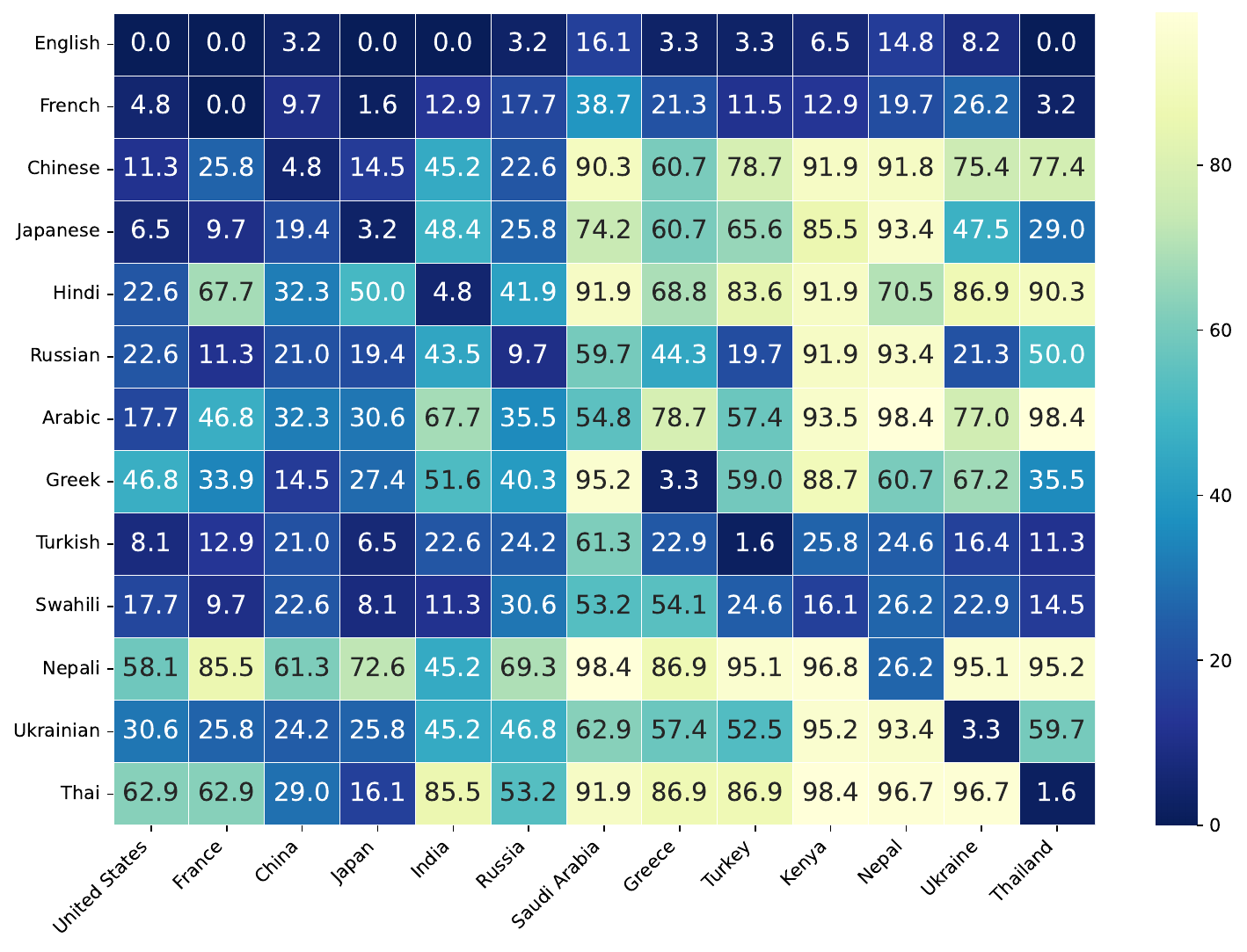}  
    \caption{Country-Specific Factual Error Rates in each language for \gemmains{2B}}
    \label{fig:factual_recall_gemma_2_2}  
\end{figure}

\begin{figure}[t]  
    \centering
    \includegraphics[width=1.\linewidth]{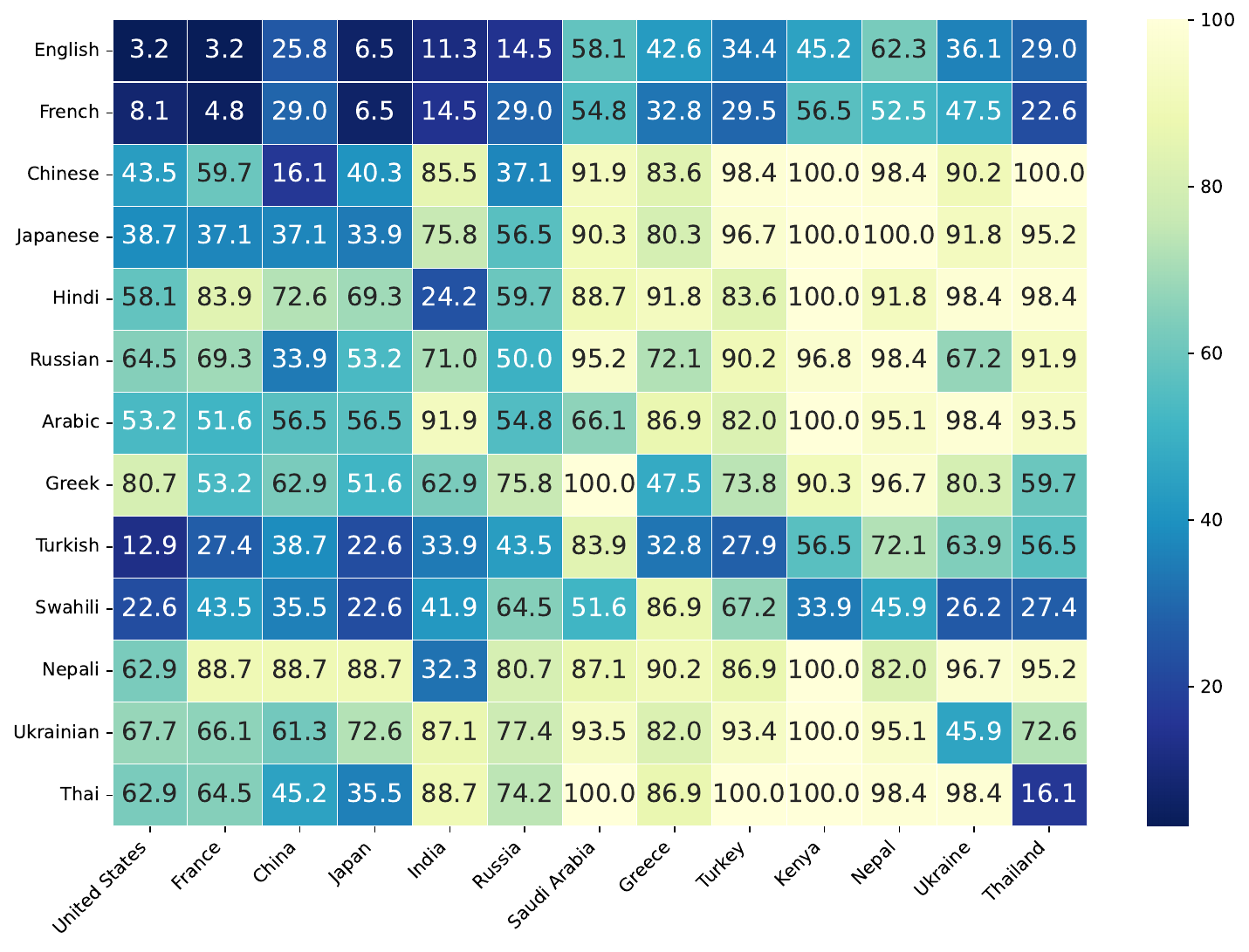}  
    \caption{Country-Specific Factual Error Rates in each language for \llamains{1B}}
    \label{fig:factual_recall_worst}  
\end{figure}

\end{document}